%% file: ODIN_AAAI.tex
\def\year{2020}\relax
\documentclass[letterpaper]{article} 
\usepackage{aaai20}  
\usepackage{times}  
\usepackage{helvet} 
\usepackage{courier}  
\usepackage[hyphens]{url}  
\usepackage{graphicx} 
\usepackage{graphicx}  
\usepackage{bm}
\usepackage{tikz}
\usepackage{amsmath}
\usepackage{mathtools}
\usepackage{algorithm,algpseudocode}
\usepackage{subcaption}
\usepackage{pgfplots}

\usetikzlibrary{calc}
\tikzset{mynode/.style={draw,circle, minimum size = 0.7cm}}
\usetikzlibrary{positioning,shapes,arrows}

\usetikzlibrary{external}
\newcommand{\citet}[1]{\citeauthor{#1} \shortcite{#1}}
\newcommand{\citep}{\cite}

\title{ODIN: ODE-Informed Regression for Parameter and State Inference in Time-Continuous Dynamical Systems}
\author{
Philippe Wenk,\thanks{Equal contribution}\textsuperscript{\rm 1,2}
Gabriele Abbati,$^*$\textsuperscript{\rm 3}
Michael A Osborne,\textsuperscript{\rm 3}\\
\Large \textbf{Bernhard Sch\"{o}lkopf,\textsuperscript{\rm 4}
Andreas Krause,\textsuperscript{\rm 1}
Stefan Bauer\textsuperscript{\rm 4}}\\
\textsuperscript{\rm 1}Learning and Adaptive Systems Group, ETH Z\"{u}rich\\ 
\textsuperscript{\rm 2}Max Planck ETH Center for Learning Systems\\
\textsuperscript{\rm 3}Department of Engineering Science, University of Oxford\\
\textsuperscript{\rm 4}Empirical Inference Group, Max Planck Institute for Intelligent Systems\\
Correspondence to: \texttt{wenkph@ethz.ch}, \texttt{gabb@robots.ox.ac.uk}\\
}
\begin{document}

	\maketitle
	
	\newtheorem{theorem}{Theorem}
	\newtheorem{example}[theorem]{Example}
	\newtheorem{remark}[theorem]{Remark}
	\newtheorem{corollary}[theorem]{Corollary}
	\newtheorem{definition}[theorem]{Definition}
	\newtheorem{lemma}[theorem]{Lemma}
	\newtheorem{proposition}[theorem]{Proposition}
	
	\begin{abstract}
		Parameter inference in ordinary differential equations is an important problem in many applied sciences and in engineering, especially in a data-scarce setting. In this work, we introduce a novel generative modeling approach based on constrained Gaussian processes and leverage it to build a computationally and data efficient algorithm for state and parameter inference. In an extensive set of experiments, our approach outperforms the current state of the art for parameter inference both in terms of accuracy and computational cost. It also shows promising results for the much more challenging problem of model selection.
	\end{abstract}
	
	\section{Introduction}
	\input{sections/intro.tex}
	\section{Background}
	\input{sections/background.tex}
	\section{ODE-Informed Regression}
	\input{sections/theory.tex}
	\section{Experiments}
	\input{sections/experiments.tex}
	\section{Discussion}
	\input{sections/discussions.tex}
	
	\subsubsection{Acknowledgments.}
	This research was supported by the Max Planck ETH Center for Learning Systems. GA acknowledges funding from Google DeepMind and University of Oxford. This project has received funding from the European Research Council (ERC) under the European Union’s Horizon 2020 research and innovation programme grant agreement No 815943.
	
	\bibliographystyle{aaai}
	\bibliography{references.bib}
	
	\appendix
	\onecolumn
	\section{Supplementary Material}
	\input{sections/appendix.tex}

\end{document}

%% file: sections/intro.tex
Ordinary differential equations (ODEs) represent a ubiquitous tool for modeling problems in many quantitative sciences and engineering. While with first principles and expert knowledge it is often possible to work out a parametric form for the equations that model a system of interest, in most cases there is no closed-form solution, making parameter identification problematic. One needs to rely on numerical schemes, which can be sub-optimal given that the exact system trajectory is usually unknown: typically, we observe noisy measurements of the true trajectory only at some discrete time points. This problem has been extensively studied in the past with classical approaches based on numerical integration (e.g. \citet{bard1974nonlinear} and \citet{benson1979parameter}), also in the relevant context of model selection \citep{chen2017network,wu2014sparse}.

Classical approaches iteratively propose new sets of parameters and then evaluate them by numerical integration: the estimated trajectory can then be compared against the observed data. Among others, \citet{varah1982spline} argue that this procedure can be turned on its head to improve computational performance. In principle, finding good parameters is equivalent to denoising the states, since adequate ODE parameters will lead to a trajectory that is close to ground truth. In particular, \citet{varah1982spline} first fit a spline curve to the observations to approximate the true trajectory, and subsequently match state and derivative estimates of said splines to obtain the ODE parameters. This idea gave rise to a class of \emph{gradient matching} algorithms that rely on spline regression, kernel regression and, in a Bayesian setting, Gaussian process regression (GPR).

Gaussian processes (GPs) provide a very natural and theoretically appealing way to smooth time series, especially because they are very closely related to Kalman filtering \citep{hartikainen2010kalman}. Thus, there has been significant interest in incorporating them into the gradient matching framework, starting from the pioneering theoretical work of \citet{calderhead2009accelerating}: they propose a GP-based probabilistic modeling scheme on which they perform inference using MCMC. \citet{dondelinger2013ode} change this probabilistic setup to achieve a more efficient MCMC sampling procedure ($\mathrm{AGM}$ - Adaptive Gradient matching), while \citet{gorbach2017scalable} introduce a computationally efficient inference scheme based on variational inference (VGM - Variational Gradient Matching). Crucially, all these methods rely on a product of experts (PoE) heuristic (an alternative approach is formulated by \citet{barber2014gaussian}, later questioned by \citet{macdonald2015controversy}). However, \citet{wenk2018fast} show that the PoE leads to theoretical issues: indeed, in the graphical models proposed by \citet{calderhead2009accelerating}, \citet{dondelinger2013ode} and \citet{gorbach2017scalable}, the ODE parameters become statistically independent of the observations. Thus, \citet{wenk2018fast} propose a new graphical model that circumvents this issue and present an efficient MCMC-based inference scheme ($\mathrm{FGPGM}$ - Fast Gaussian Process based Gradient Matching). A further formulation that is based on variational inference and allows for additional inequality constraints on the derivatives is provided by \citet{lorenzi2018constraining}.

Similarly to Gaussian process-based gradient matching, \citet{gonzalez2014reproducing} and \citet{niu2016fast} use kernel ridge regression in a frequentist setting ($\mathrm{RKG2}$/$\mathrm{RKG3}$ - Reproducing Kernel based Gradient Matching). Aiming directly for point estimates of the parameters, their approaches are naturally faster than alternatives that build on the use of MCMC and Gaussian processes. Nevertheless, they rely on several trade-off parameters to be tuned via cross-validation, which can turn out to be impractical to data-scarce environments.

In our work, we extend and blend both Bayesian and frequentist viewpoints to obtain a computationally efficient algorithm that can learn states and parameters in a low-data setting. In particular, we
\begin{itemize}
	\item Present a novel generative model, rephrasing the parameter inference problem as constrained Gaussian process regression;
	\item Provide a data-efficient algorithm that concurrently estimates states and parameters;
	\item Show how all hyperparameters can be learned from data and how they can be used as an indicator for model mismatch;
	\item Provide an efficient Python implementation for public use, with a publicly available code base at \url{https://github.com/gabb7/ODIN}
\end{itemize}

%% file: sections/background.tex
	\subsection{Problem Setting}

Throughout this work, we consider $K$-dimensional dynamical systems whose evolution is described by a set of differential equations parameterized by a vector $\bm{\theta}$, i.e.
\begin{equation}
\dot{\mathbf{x}}(t) = \mathbf{f}(\mathbf{x}(t), \bm{\theta}).
\label{eq:ODE}
\end{equation}
The system is observed under additive, zero-mean Gaussian noise at $N$ discrete time points $\mathbf{t}=[t_1, \dots, t_N]$. We assume the standard deviation of the noise to be constant over time but it may differ for each state dimension. The noise is further assumed to be uncorrelated across dimensions and time points. This leads to the following observation model:
\begin{equation}
y_k(t_i) = x_k(t_i) + \epsilon_k(t_i), \hspace{.5cm} \epsilon_k(t_i) \sim \mathcal{N}(0, \sigma_k).
\label{eq:ErrorModel}
\end{equation}
Previous research efforts propose to model $\mathbf{f}$ itself as a Gaussian process (e.g. \citet{heinonen2018learning}) or to extend the framework to SDEs (e.g. \citet{ryder2018black}, \citet{abbati2019ares}). While it might be interesting to investigate how these frameworks could be combined with $\operatorname{ODIN}$, this falls outside the scope of this paper. We thus restrict ourselves explicitly to the case where we have deterministic differential equations with known parametric form.

\subsection{Temporal Regression}

In the context of dynamical systems, Gaussian processes are employed mostly to model directly the system dynamics (i.e. the function $\mathbf{f}$). GP-based gradient matching approaches the problem differently. Here GPs model the states $\mathbf{x}$ and provide an approximation for the function $\mathbf{x}$ that maps a time point $t_i$ to the corresponding state vector $\mathbf{x}(t_i)$. For the sake of readability, we assume $K=1$ and denote the values of $x(t)$ stacked across time points as $\mathbf{x} = [x(t_1), \dots, x(t_N)]$. As shown in the experiment section, the extension to $K>1$ is straightforward: $K$ independent Gaussian processes can be stacked to model each state independently.

While our method can theoretically work with any nonlinear, differentiable regression technique, we choose to use Gaussian processes. GPs have superb analytical properties \citep{rasmussenGPBible} and they recently showed remarkable empirical results in the context of parameter inference for ODEs \citep{lorenzi2018constraining,wenk2018fast}. Moreover, thanks to the representer theorem \citep{scholkopf2001generalized}, Gaussian processes are closely related to kernel ridge regression: this connects GP-based gradient matching approaches to the reproducing-kernel-based ones (e.g. \citet{niu2016fast}).

\subsection{Gaussian Process Regression}

\begin{figure}
	\centering
	\includegraphics{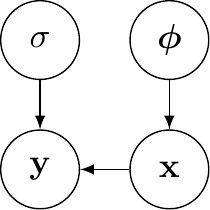}
	\caption{Generative model for standard Gaussian process regression. Given kernel hyperparameters $\bm{\phi}$ and observation noise standard deviation $\sigma$, the probability densities for the states $\mathbf{x}$ and their noisy observations $\mathbf{y}$ are fully determined.}
	\label{fig:GPRegression}
	\ \\
	\includegraphics{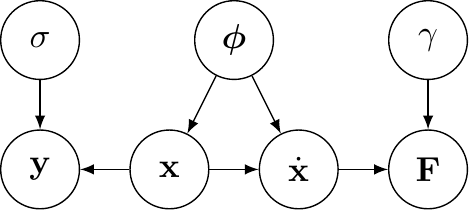}
	\caption{Generative model for GP regression with derivative observations $\mathbf{F}$, for which we use a Gaussian observation model with variance $\gamma$ (as in $\operatorname{ODIN}$). Due to the GP prior, $\mathbf{x}$ and $\dot{\mathbf{x}}$ are jointly Gaussian with known probability densities once the kernel hyperparameters $\bm{\phi}$ are determined.}
	\label{fig:GPWithDerivs}
\end{figure}

As in standard GP regression, we start by choosing a covariance function (or kernel) $k_{\bm{\phi}}$, which is parameterized by a set of hyperparameters $\bm{\phi}$. The kernel is used to compute a covariance matrix $\mathbf{C}_{\bm{\phi}}$, whose elements are given by $\left[\mathbf{C}_{\bm{\phi}}\right]_{i, j} = k_{\bm{\phi}}(t_i, t_j)$. $\mathbf{C}_{\bm{\phi}}$ can be used to define a zero-mean prior over the true states $\mathbf{x}$ at the observation times $\mathbf{t}$:
\begin{equation}
p(\mathbf{x} \mid \bm{\phi}) = \mathcal{N}(\mathbf{x} \mid \mathbf{0}, \mathbf{C}_{\bm{\phi}}).
\end{equation}
The noise model from Equation \eqref{eq:ErrorModel} yields a Gaussian likelihood for the observations $\mathbf{y}$:
\begin{equation}
p(\mathbf{y} \mid \mathbf{x}, \sigma) = \mathcal{N}(\mathbf{y} | \mathbf{x}, \sigma^2 \mathbf{I}).
\end{equation}
Using Bayes rule and observing the fact that a product of two Gaussians in the same variables is again a Gaussian, we obtain the classic GP posterior
\begin{equation}
p(\mathbf{x} \mid \mathbf{y}, \sigma, \bm{\phi}) = \mathcal{N}(\mathbf{x} \mid \bm{\mu}_{\mathbf{x}}, \bm{\Sigma}_{\mathbf{x}}),
\label{eq:statePosterior}
\end{equation}
\begin{alignat}{2}
&\text{where} \qquad \qquad &\bm{\mu}_{\mathbf{x}} &= \mathbf{C}_{\bm{\phi}} (\mathbf{C}_{\bm{\phi}} + \sigma^2\mathbf{I})^{-1} \mathbf{y}\\
&\text{and} &\bm{\Sigma}_{\mathbf{x}} &= \sigma^2 (\mathbf{C}_{\bm{\phi}} + \sigma^2\mathbf{I})^{-1} \mathbf{C}_{\bm{\phi}}
\end{alignat}
A graphical representation of this generative model can be found in Figure \ref{fig:GPRegression}. Throughout this paper, we assume that $k_{\phi}$ is \textit{differentiable} w.r.t. both of its arguments.

\begin{figure*}[h]
	\centering
	\begin{subfigure}[t]{0.325\textwidth}
		\centering
		\includegraphics{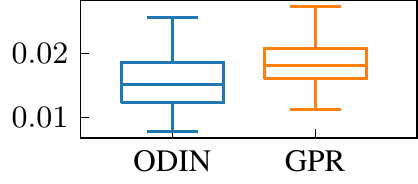}
		\caption{Lotka-Volterra}
	\end{subfigure}
	\hfill
	\begin{subfigure}[t]{0.325\textwidth}
		\centering
		\includegraphics{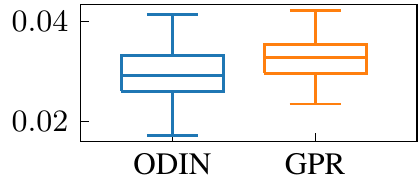}
		\caption{FitzHugh-Nagumo}
	\end{subfigure}
	\hfill
	\begin{subfigure}[t]{0.325\textwidth}
		\centering
		\includegraphics{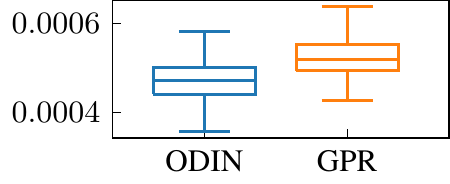}
		\caption{Protein Transduction}
	\end{subfigure}
	\caption{RMSE of state estimates using vanilla GP regression and $\operatorname{ODIN}$ on the benchmark systems (low noise case). While GPR can only access the noisy observations $\mathbf{y}$, $\operatorname{ODIN}$ considers the parametric form of $\mathbf{f}(\mathbf{x}, \bm{\theta})$ (with no information about $\bm{\theta}$). This additional regularization contributes towards more accurate estimations.}
	\label{fig:stateRMSE}
\end{figure*}

\subsection{Gaussian Process Regression with Derivatives}
As previously noted e.g. by \citet{solak2003derivative}, the estimate of the posterior distribution of the states given by Equation \eqref{eq:statePosterior} can be further refined if we consider access to noisy observations of the derivatives. Let us then assume we have additional observations $\mathbf{F}$ that are generated by $F(t_i) = \dot{x}(t_i) + \delta(t_i)$, where $\delta(t_i) \sim \mathcal{N}(0, \gamma)$.
Incorporating such derivatives is straightforward: since Gaussian processes are closed under linear operations, the distribution over the derivatives is again a Gaussian process. Following the notation of \citet{wenk2018fast} and the argumentation in their appendix, we obtain
\begin{equation}
p(\dot{\mathbf{x}} \mid \mathbf{x}, \bm{\phi}) = \ \mathcal{N}(\dot{\mathbf{x}} \mid \mathbf{D} \mathbf{x}, \mathbf{A}),
\label{eq:derivPrior}
\end{equation}
where the exact form of the matrices $\mathbf{D}$ and $\mathbf{A}$ is omitted for simplicity, but can be found in the supplementary material. Equation \eqref{eq:derivPrior} can now be combined with the likelihood for the derivative observations
\begin{equation}
p(\mathbf{F} \mid \dot{\mathbf{x}}, \mathbf{\gamma}) = \mathcal{N}(\mathbf{F} \mid \dot{\mathbf{x}}, \gamma \mathbf{I}).
\end{equation} 
This leads to the generative model shown in Figure \ref{fig:GPWithDerivs}. Just like in standard Gaussian process regression, all posteriors of interest can be calculated analytically, since all probability densities are Gaussian distributions in $\mathbf{x}$, $\dot{\mathbf{x}}$ or linear transformations thereof.

%% file: sections/theory.tex
\begin{figure*}[h]
	\centering
	\begin{subfigure}[t]{0.325\textwidth}
		\centering
		\includegraphics{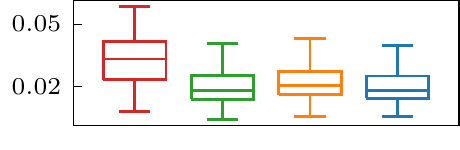}
	\end{subfigure}
	\hfill
	\begin{subfigure}[t]{0.325\textwidth}
		\centering
		\includegraphics{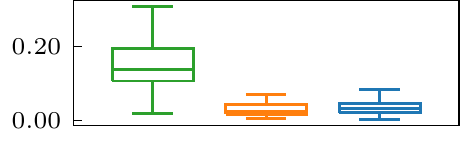}
	\end{subfigure}
	\hfill
	\begin{subfigure}[t]{0.325\textwidth}
		\centering
		\includegraphics{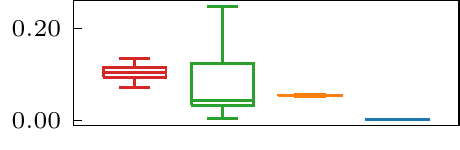}
	\end{subfigure}
	\begin{subfigure}[t]{0.325\textwidth}
		\centering
		\includegraphics{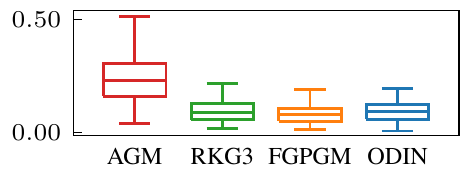}
		\caption{Lotka-Volterra}
	\end{subfigure}
	\hfill
	\begin{subfigure}[t]{0.325\textwidth}
		\centering
		\includegraphics{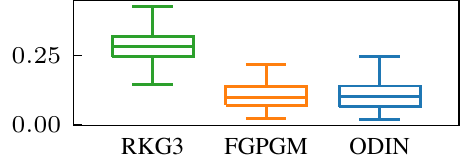}
		\caption{FitzHugh-Nagumo}
	\end{subfigure}
	\hfill
	\begin{subfigure}[t]{0.325\textwidth}
		\centering
		\includegraphics{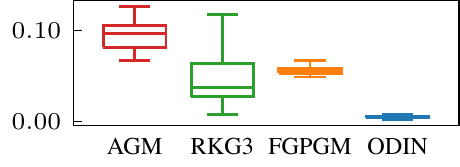}
		\caption{Protein Transduction}
	\end{subfigure}
	\caption{Trajectory RMSE for parameter inference on the benchmark systems. The top row shows the low noise case with $\sigma=0.1$ for LV, $SNR=100$ for FHN and $\sigma=0.001$ for PT. The bottom row shows the high noise case with $\sigma=0.5$ for LV, $SNR=10$ for FHN and $\sigma=0.01$ for PT.}
	\label{fig:trajectoryRMSE}
\end{figure*}

\subsection{Gaussian Process-based Gradient Matching}
Given the model in Figure \ref{fig:GPWithDerivs}, the main challenge consists in including the mathematical expressions of the ODEs in a meaningful way. In traditional GP-based gradient matching, the ODEs are introduced as a second generative model for $\mathbf{F}$ or $\dot{\mathbf{x}}$. The latter is then combined with the Gaussian process model of Figure \ref{fig:GPWithDerivs} to establish a probabilistic link between the observations $\mathbf{y}$ and the parameters $\bm{\theta}$. However, the GP model fully determines the probability densities of $\mathbf{F}$ and $\dot{\mathbf{x}}$. Thus, the two generative models have to be combined using some heuristic, like the product of experts \citep{calderhead2009accelerating,dondelinger2013ode,gorbach2017scalable} or an additional Dirac delta function forcing equality \citep{wenk2018fast}.

The resulting, unified generative model is then used to approximate the posterior of $\mathbf{x}$ and $\bm{\theta}$ through Bayesian inference techniques, e.g. MCMC \citep{calderhead2009accelerating,dondelinger2013ode,wenk2018fast} or variational mean field \citep{gorbach2017scalable}. Inference for these algorithms consists in computing mean and standard deviation of an approximate posterior to get estimates that include uncertainty. As we shall see in the experiment section, this works well for sufficiently tame dynamics and identifiable systems, but struggles to produce meaningful results for multi-modal posteriors. Crucially, practical systems often produce multi-modal posteriors and suffer from unidentifiability without strong priors (e.g. \citet{stephan2008nonlinear}, \citet{hass2017predicting}). As we shall see, ODE-informed regression can overcome this issue.

\subsection{ODIN: ODE-Informed Regression}
\label{sec:ODIN_theory}
To avoid the problems associated with the two probabilistic models of traditional GP-based gradient matching, $\operatorname{ODIN}$ does not include the ODEs via a separate generative model. Instead, they are introduced at inference time in the form of \emph{constraints}, essentially solving a constrained MAP problem. 

We start with the joint density of the Gaussian process described in Figure \ref{fig:GPWithDerivs}, denoted by $p(\mathbf{y}, \mathbf{x}, \dot{\mathbf{x}}, \mathbf{F} \mid \sigma, \gamma, \bm{\phi})$. As a result of the Gaussian observation model for $\mathbf{F}$, $\dot{\mathbf{x}}$ can be marginalized out analytically, leading to
\begin{align}
\label{eq:jointDensity}
p(&\mathbf{y}, \mathbf{x}, \mathbf{F} \mid \sigma, \gamma, \bm{\phi}) =\\
&\mathcal{N}(\mathbf{y} \mid \mathbf{x}, \sigma^2 \mathbf{I})\ 
\mathcal{N}(\mathbf{x} \mid \mathbf{0}, \mathbf{C}_{\bm{\phi}})\ 
\mathcal{N}(\mathbf{F} \mid \mathbf{D} \mathbf{x}, \mathbf{A} + \gamma \mathbf{I}). \nonumber
\end{align}
Assuming fixed values for $\sigma, \bm{\phi}$ and $\gamma$, this equation can be simplified by taking the logarithm, discarding all terms that do not explicitly depend on the states $\mathbf{x}$ and the derivative observations $\mathbf{F}$ and ignoring multiplicative factors to obtain
\begin{align}
\tilde{\mathcal{R}}(&\mathbf{x}, \mathbf{F}, \mathbf{y}) = \label{eq:ODE_risk} \\
&||\mathbf{x}||^2_{\mathbf{C}_{\bm{\phi}}^{-1}}
+ 
||\mathbf{x} - \mathbf{y}||^2_{\sigma^{-2} \mathbf{I}}
+ 
||\mathbf{F} - \mathbf{D}\mathbf{x}||^2_{(\mathbf{A} + \gamma \mathbf{I})^{-1}}, \nonumber
\end{align}
where $||\mathbf{u}||^2_\mathbf{M} \coloneqq \mathbf{u}^T \mathbf{M} \mathbf{u}$ is the norm of the vector $\mathbf{u}$ weighted by a positive-definite matrix $\mathbf{M}$.

The key mechanism behind $\operatorname{ODIN}$ lies in how we obtain values for $\mathbf{F}$. In principle, $\mathbf{F}$ could be marginalized out to recover standard GP regression. However, this is not desirable, as we would ignore the ODE information. Instead, $\operatorname{ODIN}$ includes the ODEs as additional constraints in the optimization problem: rather than keeping $\mathbf{F}$ completely flexible, we assume the existence of a parameter vector $\bm{\theta}$ that links the derivative observations to the ODEs. More formally,
\begin{align}
&\mathbf{x}, \mathbf{F} = \arg \min_{\mathbf{x}, \bm{\theta}} \tilde{\mathcal{R}}(\mathbf{x}, \mathbf{F}, \mathbf{y}) \\
\text{s. t.} \quad &\exists \bm{\theta} \quad \text{with} \quad \mathbf{f}(\mathbf{x}(t_i), \bm{\theta}) = \mathbf{F}_i \quad \text{for all }i.
\end{align}
As it turns out, these constraints can be incorporated in the optimization problem by directly substituting $\mathbf{F}$ with the corresponding contribution from the ODEs, leading to
\begin{equation}
\mathbf{x}, \bm{\theta} = \arg \min_{\mathbf{x}, \bm{\theta}} \mathcal{R}(\mathbf{x}, \bm{\theta}, \mathbf{y}), \label{eq:ODIN_Objective_Gamma_free}
\end{equation}
where $\mathcal{R}(\mathbf{x}, \bm{\theta}, \mathbf{y}) \coloneqq \tilde{\mathcal{R}}(\mathbf{x}, \bm{f}(\mathbf{x}, \bm{\theta}), \mathbf{y})$.

This constitutes the key idea behind $\operatorname{ODIN}$. Instead of providing direct observations of the derivatives, we generate them via the ODEs after fixing the ODE parameters $\bm{\theta}$.

Similarly to classical frequentist methods \citep{varah1982spline,niu2016fast}, $\mathcal{R}(\mathbf{x}, \bm{\theta}, \mathbf{y})$ punishes divergence between the states $\mathbf{x}$ and the observations $\mathbf{y}$, as well as between the output of the ODEs, $\mathbf{f}(\mathbf{x}, \bm{\theta})$, and the derivatives estimated by the regressing GP; the regularization term avoids overfitting.	However, in sharp contrast to frequentist approaches, all trade-off parameters are naturally provided by the GP framework, once the hyperparameters $\bm{\phi}$ and noise levels $\sigma$ and $\gamma$ are fixed. As we will see later, the absence of cross-validation for hyperparameter learning crucially improves accuracy in sparsely observed systems. 

\begin{figure*}[h]
	\centering
	\begin{subfigure}[t]{0.245\textwidth}
		\centering
		\includegraphics{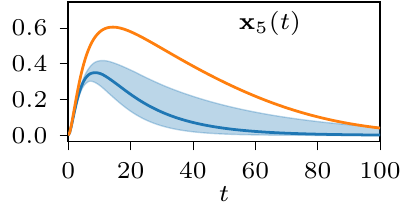}
		\caption{$\mathrm{AGM}$}
	\end{subfigure}
	\hfill
	\begin{subfigure}[t]{0.245\textwidth}
		\centering
		\includegraphics{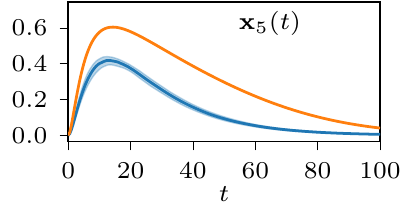}
		\caption{$\mathrm{FGPGM}$}
	\end{subfigure}
	\hfill
	\begin{subfigure}[t]{0.245\textwidth}
		\centering
		\includegraphics{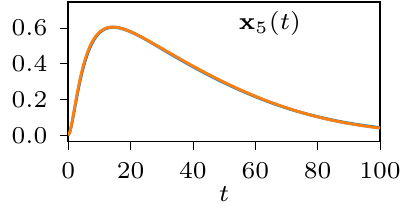}
		\caption{$\operatorname{ODIN}$}
	\end{subfigure}
	\hfill
	\begin{subfigure}[t]{0.245\textwidth}
		\centering
		\includegraphics{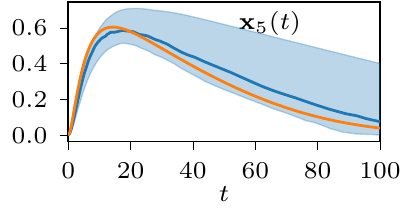}
		\caption{$\mathrm{RKG3}$}
	\end{subfigure}
	\caption{Comparison of the trajectories obtained by numerically integrating the inferred parameters of the Protein Transduction system for $\sigma=0.01$. The solid blue line is the median trajectory, while for clarity we shaded the area between the 25\% and 75\% quantiles. The orange trajectory represents the ground truth.}
	\label{fig:medianTrajectories}
\end{figure*}

\begin{algorithm}[h]
	\caption{$\operatorname{ODIN}$}
	\label{alg:ODIN_auto}
	\begin{algorithmic}[1]
		\State{\textbf{Input:}\quad  $\mathbf{y}^{(1)}, \dots, \mathbf{y}^{(K)}, \mathbf{f}(\mathbf{x}, \boldsymbol{\theta})$}
		\State{\emph{Step 1: State-independent GP regression}}
		\ForAll{$ k \in K$}
		\State{Standardize time $\mathbf{t}$ and observations $\mathbf{y}_k$.}
		\State{Fit $\bm{\phi_k}$ and $\sigma_k$ using empirical Bayes, i.e. maximize \color{white} ..... \color{black}$p(\mathbf{y}^{(k)} | \mathbf{t}, \boldsymbol{\phi}_k, \sigma_k)$.}
		\State{Initialize $\mathbf{x}_k$ using the mean $\bm{\mu}_k$ of the trained GP.}
		\EndFor
		\State{\emph{Step 2: ODE Information Incorporation}}
		\State{Initialize $\bm{\theta}$ randomly.}
		\State{Initialize $\gamma_1, \dots, \gamma_K = 1.0$}
		\State{Apply L-BFGS-B to solve the optimization problem \eqref{eq:startAutoRisk} and obtain $\hat{\mathbf{x}}, \hat{\bm{\theta}}, \hat{\gamma}_1, \dots, \hat{\gamma}_K.$}
		\State{\textbf{Return:} $\hat{\mathbf{x}}, \hat{\bm{\theta}}, \hat{\gamma}_1, \dots, \hat{\gamma}_K$}
	\end{algorithmic}
\end{algorithm}

\subsection{Derivative Observation Model}
Let us recall that we conceptually substitute the ODE outputs with derivative observations that are subjected to Gaussian noise, with variance $\gamma$. In this process, we can interpret in an intuitive but meaningful way: during and after training,
the ODE outputs and the GP derivative estimates can deviate from the ground truth and thus differ from each other. This divergence is accounted for by $\gamma$. In classical GP-based approaches \citep{dondelinger2013ode,gorbach2017scalable,wenk2018fast}, $\gamma$ is treated as a random variable whose values are independent of the inference procedure; sometimes it is fixed a priori \citep{gorbach2017scalable,wenk2018fast}. However, we can expect that the divergence between ODEs and GP derivatives would be larger in the early steps of training, while it should decrease when the ODEs describe well to the ground truth. Thus, it is sensible to automatically adapt $\gamma$ to reflect the current quality of the estimates.

The $\operatorname{ODIN}$ framework can be adjusted to reflect this reasoning. To obtain Equation \eqref{eq:ODE_risk}, we implicitly assumed $\gamma$ to be constant. If we rather treat it as an optimization parameter, the objective of Equation \eqref{eq:ODIN_Objective_Gamma_free} changes to
\begin{equation}
\mathbf{x}, \bm{\theta}, \gamma = \arg \min_{\mathbf{x}, \bm{\theta}, \gamma} \mathcal{R}(\mathbf{x}, \bm{\theta}, \mathbf{y}, \gamma) \label{eq:startAutoRisk}
\end{equation}
where
\begin{equation}
\mathcal{R}(\mathbf{x}, \bm{\theta}, \mathbf{y}, \gamma) = \mathcal{R}(\mathbf{x}, \bm{\theta}, \mathbf{y}) +  \log(\det(\mathbf{A} + \gamma \mathbf{I})).
\label{eq:FinalRiskGamma}
\end{equation}

If $\gamma$ is part of the optimization procedure, the contribution of the normalization constant in Equation \eqref{eq:jointDensity} can not be ignored when deriving the risk appearing in Equation \eqref{eq:FinalRiskGamma}. In practice, similarly to the log-determinant in standard GP regression, this term acts as an Occam's razor by preventing an excessive growth of $\gamma$ if the GP derivatives and the ODE outputs differ significantly.

The final $\operatorname{ODIN}$ routine is summarized as Algorithm \ref{alg:ODIN_auto}.

\subsection{Remarks}

Throughout this work, we assume to have access to observations $\mathbf{y}$ that are subjected to the noise model described in Equation \eqref{eq:ErrorModel}. However, the Gaussian noise assumption is only needed when deriving the term $||\mathbf{x} - \mathbf{y}||^2_{\sigma^{-2} \mathbf{I}}$ in $\mathcal{R}$. Thus, it could be straightforward to accommodate for alternative noise models by adjusting the corresponding term in the risk formula. On the other side, a Gaussian noise model (with variance $\gamma$) is a strict requirement for the derivative observations, as it is necessary to marginalize $\dot{\mathbf{x}}$ analytically.

As demonstrates in the experiments section, in the case of perfect ODEs (i.e. when we know the true parametric form a priori), $\gamma$ can in principle be set to zero; nevertheless, when that is not the case it provides an effective mechanism for detecting model mismatch and helps with the challenging problem of model selection.

%% file: sections/experiments.tex
\begin{figure*}[h]
	\centering
	\begin{subfigure}[t]{0.24\textwidth}
		\includegraphics{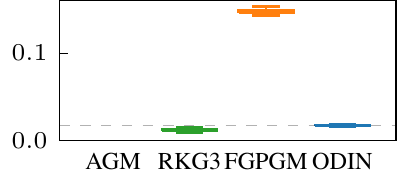}
		\caption{$\theta_5$, $\sigma=0.001$}
	\end{subfigure}
	\hfill
	\begin{subfigure}[t]{0.24\textwidth}
		\centering
		\includegraphics{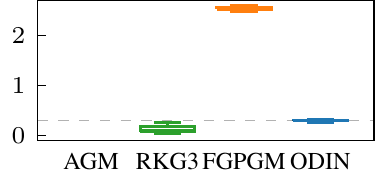}
		\caption{$\theta_6$, $\sigma=0.001$}
	\end{subfigure}
	\hfill
	\begin{subfigure}[t]{0.24\textwidth}
		\centering
		\includegraphics{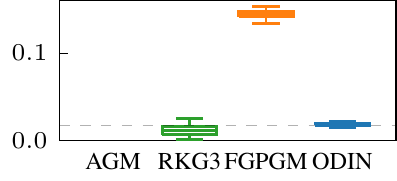}
		\caption{$\theta_5$, $\sigma=0.01$}
	\end{subfigure}
	\hfill
	\begin{subfigure}[t]{0.24\textwidth}
		\centering
		\includegraphics{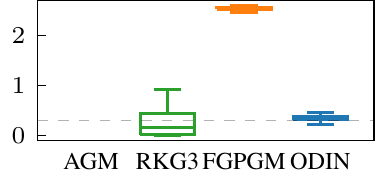}
		\caption{$\theta_6$, $\sigma=0.01$}
	\end{subfigure}
	\caption{Parameter estimates for Protein Transduction for $\sigma=0.001$ (a-b) and $\sigma=0.01$ (c-d). Showing median, 50\% and 75\% quantiles over 100 independent noise realizations. The dashed line indicates the ground truth.}
	\label{fig:identifiability}
\end{figure*}

In this section, we demonstrate the versatility of $\operatorname{ODIN}$ and compare its performance to various state-of-the-art methods. We start by comparing its parameter inference capabilities to various state-of-the-art inference schemes on three commonly used benchmark systems: the Lotka-Volterra (LV) predator-prey model \citep{lotka1932growth}; the FitzHugh-Nagumo (FHN) neuronal model \citep{fitzhugh1961impulses,nagumo1962active}; the chemical protein transduction (PT) system as seen in \citet{vyshemirsky2007bayesian}. All three systems have already been studied extensively in the context of gradient matching \citep{calderhead2009accelerating,dondelinger2013ode,gorbach2017scalable,wenk2018fast} and thus represent a clear benchmark. For completeness, we restate the concrete parametric form in the supplementary material, together with the ground truth for all parameters. In addition to state and parameter inference, we show how $\operatorname{ODIN}$ can be used for model selection, a missing feature for every comparison method here considered. Finally, we prove linear scaling behavior of $\operatorname{ODIN}$ in the state dimension $K$ by investigating its performance on a high-dimensional, fourth benchmark system with up to 1000 states.

\subsection{Evaluation Details and Data Creation}
All experimental datasets are generated using numerical simulations. Thus, the ground truth for both the states $\mathbf{x^*}$ and parameters $\bm{\theta}^*$ is always available. Following \citet{wenk2018fast}, we employ the trajectory RMSE as a metric to compare the quality of parameter estimates. For ease of reference, we restate the definition in Definition \ref{def:trajRMSE}.

\begin{definition}[Trajectory RMSE]
	\label{def:trajRMSE}
	Let $\hat{\bm{\theta}}$ be the parameters estimated by an inference algorithm. Let $\mathbf{t}$ be the vector collecting the observation times. Define $\tilde{x}(t)$ as the trajectory one obtains by integrating the ODEs using the estimated parameters, but the true initial value, i.e.
	\begin{align}
	\tilde{x}(0) &= \mathbf{x}^*(0)\\
	\tilde{x}(t) &= \int_{0}^{t}f(\tilde{x}(s), \hat{\bm{\theta}}) ds
	\end{align}
	and define $\tilde{\mathbf{x}}$ element-wise as its evaluation at observation times $\mathbf{t}$, i.e. $\tilde{x}_i = \tilde{x}(t_i)$. The trajectory RMSE is then defined as 
	\begin{equation}
	\textrm{tRMSE} \coloneqq \frac{1}{N}||\tilde{\mathbf{x}} - \mathbf{x}||_2,
	\end{equation}
	where $||.||_2$ denotes the standard Euclidean 2-norm.
\end{definition}

To evaluate the robustness of each algorithm w.r.t. different observation noise realizations, we always run 100 repetitions for every experimental setting. In each repetition, we keep $\mathbf{x}^*$ and $\bm{\theta}^*$ fixed and only sample the noise on $\mathbf{y}$. Results are then reported as quantiles over these 100 runs.

\begin{table*}[h]
	\caption{Median and standard deviation of computation time (in seconds) for parameter inference over 100 independent noise realizations.}
	\label{tab:runTimes}
	\centering
	\scalebox{1}{
		\begin{tabular}{ | l | c | c | c | c |}
			\hline
			& $\mathrm{AGM} [s]$ & $\mathrm{RKG3} [s]$ & $\mathrm{FGPGM} [s]$ & $\operatorname{ODIN} [s]$ \\
			\hline
			LV, $\sigma = 0.1$ & $4548.0 \pm 453.8$ & $79.0 \pm 19.0$ & $3169.5 \pm 90.1$ & $\mathbf{13.4 \pm 5.1}$  \\
			LV, $\sigma = 0.5$ & $4545.0 \pm 558.5$ & $76.5 \pm 15.8$ & $3187.5 \pm 340.9$ & $\mathbf{11.4 \pm 5.1}$  \\
			\hline
			FHN, $SNR = 100$ & / & $74.5 \pm 14.3$ & $8678.0 \pm 482.7$ & $\mathbf{5.8 \pm 3.5}$  \\
			FHN, $SNR = 10$ & / & $77.5 \pm 12.3$ & $8677.0 \pm 487.8$ & $\mathbf{4.4 \pm 3.8}$  \\
			\hline
			PT, $\sigma = 0.001$ & $29776.5 \pm 4804.7$ & $469.0 \pm 21.6$ & $20291.5 \pm 435.3$ & $\mathbf{8.9 \pm 1.5}$  \\
			PT, $\sigma = 0.01$ & $30493.0 \pm 1470.4$ & $480.0 \pm 42.0$ & $20437.0 \pm 713.2$ & $\mathbf{20.6 \pm 3.75}$  \\
			\hline 
	\end{tabular}}
\end{table*}

\subsection{State and Parameter Inference}
\label{subsec:state_par_inf}
In the parameter inference setting, the true parametric form of the dynamical system is assumed to be provided by a practitioner, derived through first principles or expert knowledge. Thus, together with the noisy observations $\mathbf{y}$, we have access to the true parametric form $\dot{\mathbf{x}} = f(\mathbf{x}, \bm{\theta})$. The goal is to recover the true states $\mathbf{x}$ and parameters $\bm{\theta}^*$ at observation time. While smoothing is important, estimating $\bm{\theta}^*$ is of greater practical importance.

Out of the three comparison algorithms we chose, $\mathrm{AGM}$~\citep{dondelinger2013ode} and $\mathrm{FGPGM}$~\citep{wenk2018fast} rely on Gaussian processes and MCMC inference, while $\mathrm{RKG3}$~\citep{niu2016fast} chooses a frequentist, kernel-regression-based approach. For all comparisons, implementations provided by the respective authors are used. Once more in accordance to the gradient matching literature, evaluations include both a low and a high noise setting for every system. 

As shown by \citet{solak2003derivative}, including direct observations of $\mathbf{F}$ can improve the accuracy of GP regression. $\operatorname{ODIN}$ does not have access to such observations, but it leverages the parametric form of the ODEs as a regularizer when performing state inference. As can be seen in Figure \ref{fig:stateRMSE}, this regularization actually improves the estimates of the states. This fact motivates a key difference to \citep{calderhead2009accelerating}, who propose to first fit the states using GPR and then perform gradient matching while keeping the states fixed. In the following, we show how $\operatorname{ODIN}$ can learn reliable parameters, improving the current state of the art in terms of accuracy and run time.

\subsubsection{Accuracy}
In Figure \ref{fig:trajectoryRMSE} we compare the trajectory RMSE for the three benchmark systems. While the total tRMSE is an effective indicator for the overall performance, we also include the state-wise tRMSE in the supplementary material. Unfortunately, $\mathrm{AGM}$ was unstable on FitzHugh-Nagumo despite serious hyper-prior tuning efforts on our side. We thus do not have any results for this case. To help visualizing the raw numbers obtained by the tRMSE, we also report in Figure \ref{fig:medianTrajectories} the trajectories obtained by numerically integrating the inferred parameters. While here we report only one state for the high noise case of Protein Transduction, a full set of plots can be found in the supplement.

\subsubsection{Run time}
In Table \ref{tab:runTimes}, we list the median training times (in seconds) and the corresponding standard deviation of all algorithms on the three parameter inference benchmark systems. It is evident (and not unexpected) that the optimization-based algorithms $\operatorname{ODIN}$ and $\mathrm{RKG3}$ are orders of magnitude faster than the MCMC-based $\mathrm{FGPGM}$ and $\mathrm{AGM}$. Furthermore, the need for cross-validation schemes in $\mathrm{RKG3}$ seems to increase its run time roughly by an order of magnitude when compared to $\operatorname{ODIN}$.

\subsubsection{Identifiability}
\begin{figure}[h]
	\centering
	\begin{subfigure}[t]{0.23\textwidth}
		\includegraphics{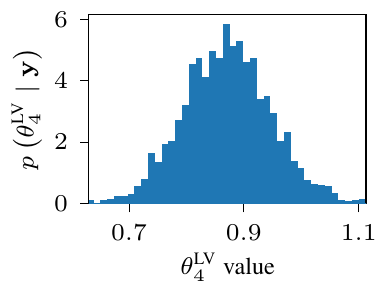}
	\end{subfigure}
	\hfill
	\begin{subfigure}[t]{0.23\textwidth}
		\includegraphics{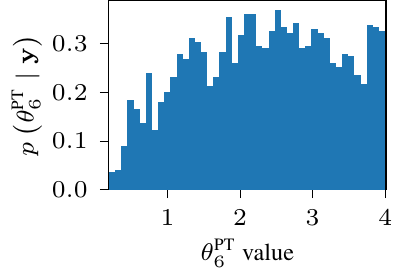}
	\end{subfigure}
	\caption{Parameter marginal distributions of $\theta_4$ of Lotka-Volterra and $\theta_6$ of Protein Transduction for one sample rollout with fixed $\gamma$. While the LV marginal is nicely Gaussian, the PT marginal is much wilder.}
	\label{fig:marginals}
\end{figure}

While both LV as well as FHN models are relatively simple, Protein Transduction (PT) still represents a considerable challenge. Amongst others, both \citet{dondelinger2013ode} and \citet{wenk2018fast} claim that the two parameters $\theta_5$ and $\theta_6$ are only weakly identifiable. However, a quick experiment with different numerical values for those ODE parameters shows that they are actually identifiable. Indeed, neither $\mathrm{RKG3}$ and $\operatorname{ODIN}$ seem to suffer from identifiability problems.
For $\operatorname{ODIN}$, this can be attributed to two key differences, the inference scheme and the flexible $\gamma$. Both $\mathrm{AGM}$ and $\mathrm{FGPGM}$ ultimately return the posterior mean of the parameter marginals. In Figure \ref{fig:marginals}, we show example marginals for fixed $\gamma$. While these distributions are Gaussian-shaped for Lotka-Volterra, they are much wilder for PT. If we were to keep the $\gamma$ fixed, $\operatorname{ODIN}$ would converge to an optimum instead of an expectation, which might be more appropriate in a multi-modal setting. However, $\operatorname{ODIN}$ does not keep $\gamma$ fixed. Instead, its $\gamma$ evolves during optimization according to the quality of the current parameters estimation, leading to an overall smoother inference. Consequently, the final parameter estimates are significantly more accurate (see Figure \ref{fig:identifiability}). For $\mathrm{AGM}$, while the ratio between $\theta_5$ and $\theta_6$ is fairly stable and reasonably not far from the correct number, the absolute parameter values have median magnitudes of roughly $10^{12}$: thus they do not appear in this figure.

\begin{table*}[!h]
	\caption{Median and standard deviation of $\gamma$ for different model misspecifications of the Lotka-Volterra system and 100 independent noise realizations.}
	\label{tab:Gamma}
	\centering
	\scalebox{1}{
		\begin{tabular}{| l | c | c | c | c |}
			\hline
			& $\mathcal{M}_{1, 1}$ & $\mathcal{M}_{0, 1}$ & $\mathcal{M}_{1, 0}$ & $\mathcal{M}_{0, 0}$ \\
			\hline
			$\gamma_1$ & $10^{-6} \pm 0.00$ & $3.01 \pm 0.23$ & $10^{-6} \pm 0.00$ & $3.03 \pm 0.24$ \\
			$\gamma_2$ & $10^{-6} \pm 0.04$ & $10^{-6} \pm 0.00$ & $1.51 \pm 0.31$ & $1.53 \pm 0.35$ \\ 
			\hline
	\end{tabular}}
\end{table*}

\subsubsection{Robustness}
Besides accurate parameter estimates, $\operatorname{ODIN}$ also exhibits more contained variance, especially compared to $\mathrm{AGM}$ and $\mathrm{RKG3}$. This is a direct consequence of the underlying GP structure, which enables efficient and stable calculation of all parameters. Furthermore, a flexible $\gamma$ seems to smooth out the optimization surface, avoiding the rugged landscapes reported by \citet{dondelinger2013ode}.

\subsubsection{Priors}
While in a Bayesian inference setting it is common to introduce a prior over $\bm{\theta}$, our graphical model in Figure \ref{fig:GPWithDerivs} does not treat $\bm{\theta}$ as a random variable. In a practical setting, we might not even know the parametric form of the ODEs: it thus seems quite difficult to justify the use of a prior. However, it should be noted that our framework can easily accommodate any prior without major modifications. An additional factor $p(\bm{\theta})$ in Equation \eqref{eq:jointDensity} directly leads to an additional summand $-\log(p(\bm{\theta}))$ in Equation \eqref{eq:FinalRiskGamma}. From a frequentist perspective, this could be interpreted as an additional regularizer, similarly to LASSO or ridge regression. Since all other summands in Equation \eqref{eq:FinalRiskGamma} grow linearly with the amount of observations $N$ and the prior contribution stays constant, the regularization term would eventually have minor influence in an asymptotic setting.

\subsection{Model Selection}
\label{sec:ModelSel}
In practice, domain experts might not be able to provide one single true model. Instead, they might indicate a set of plausible models that they would like to test against the observed data. In this section we investigate this problem, known as model selection. For empirical evaluation, we use the Lotka-Volterra system as ground truth to simulate our empirical data. We then create four different candidate models via the following two additional ODEs
\begin{alignat}{2}
\dot{x}_1(t) &= \enskip && \theta_1 x_1^2(t) + \theta_2 x_2(t) \label{eq:LV_wrong_first}, \\
\dot{x}_2(t) &= \enskip-&& \theta_3 x_2(t) \label{eq:LV_wrong_second}.
\end{alignat}
Each model is indexed as $\mathcal{M}_{i, j}$, where $i, j \in \{0, 1\}$. Here, $i=0$ indicates that the wrong equation (i.e. \ref{eq:LV_wrong_first}) is used to model the dynamics of the first state, while if $i=1$ we provide the true parametric form in that specific candidate model. In a similar fashion, $j=0$ indicates that the wrong equation (i.e. \ref{eq:LV_wrong_second}) is used to model the dynamics of the second state, otherwise $j=1$. $\operatorname{ODIN}$ is run independently for each $\mathcal{M}_{i, j}$. Besides state and parameter estimates, we thus obtain final values for $\gamma$, which are presented in Table \ref{tab:Gamma}. For numerical stability, $\gamma$ was lower bounded to $10^{-6}$ in all experiments. For the correct model $\mathcal{M}_{1, 1}$, $\gamma$ settles at this lower bound, while it converges to a much larger value in case a wrong model is used. This justifies the intuitive interpretation of $\gamma$ as a mean to account for model mismatch between the GP regressor and the ODE model. This last result proves that $\gamma$ is indeed an efficient tool for identifying true parametric forms. Interestingly, this also works dimension-wise for the mixed models $\mathcal{M}_{0, 1}$ and $\mathcal{M}_{1, 0}$, even though the states $x_1$ and $x_2$ are coupled via wrong ODEs. This can be explained by the GP regressor prioritizing states $\mathbf{x}$ close to the observations $\mathbf{y}$. Indeed, while incorrect ODEs might deteriorate the accuracy of the state estimates with wrong regularization, their detrimental effects are limited by the observation-dependent partial objective, effectively decoupling the model mismatch across dimensions.

\subsection{Linear Scaling in State Dimension}
\label{subsec:scaling}

\begin{figure}[h]
	\centering
	\includegraphics{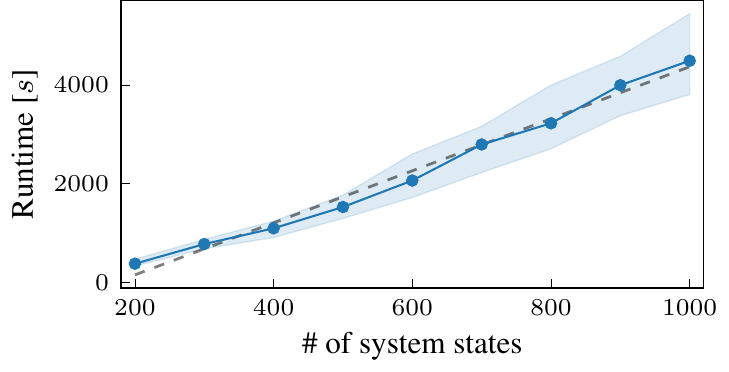}
	\caption{Run time for parameter inference on Lorenz '96 for different state dimension, with a (dashed) linear regressor fitted to the data. For each system size, we report the mean (dots) +- one standard deviation (shaded area) over 100 independent noise realizations.}
	\label{fig:scaling}
\end{figure}

A key feature of gradient matching algorithms is the linear scaling in the state dimension $K$. Following \citet{gorbach2017scalable}, we demonstrate this for $\operatorname{ODIN}$ by using the Lorenz '96 system with $\theta=8$, using 50 observations equally spaced over $t = [0, 5]$. The results are shown in Figure \ref{fig:scaling}, including a linear regressor fitted to the means with least squares.

%% file: sections/discussions.tex
Parametric ODE systems are at the backbone of many practical applications, settings where Gaussian processes and kernel regression have shown to be efficient inference tools. In this paper, we demonstrate how to combine the advantages of both approaches by using theoretical insights to extend standard GP regression. The resulting algorithm, $\operatorname{ODIN}$, significantly improves the current state of the art in terms of accuracy and runtime for parameter inference tasks and provides an appealing framework for model selection. Unlike other methods, $\operatorname{ODIN}$ does not require hyperparameter tuning and represents an out-of-the-box applicable tool for parameter inference and model selection for parametric ODE models.

%% file: sections/appendix.tex
\subsection{GP Derivatives}
For completeness, we restate the parametric form of the GP derivative mean and covariance as introduced by \citet{calderhead2009accelerating}.
\begin{equation}
p(\dot{\mathbf{x}} \mid \mathbf{x}, \bm{\phi}) = \ \mathcal{N}(\dot{\mathbf{x}} \mid \mathbf{D} \mathbf{x}, \mathbf{A}),
\end{equation}
where
\begin{align}
\mathbf{D} &\coloneqq {'\mathbf{C}_{\bm{\phi}}} \mathbf{C}_{\bm{\phi}}^{-1},\\
\mathbf{A} &\coloneqq \mathbf{C}_{\bm{\phi}}'' - {'\mathbf{C}_{\bm{\phi}}} \mathbf{C}_{\bm{\phi}}^{-1} \mathbf{C}_{\bm{\phi}}'
\end{align}
and
\begin{align}
\left['\mathbf{C}_{\bm{\phi}} \right]_{i,j} &\coloneqq \frac{\partial}{\partial a} k_{\bm{\phi}}(a, b) \rvert_{a=t_i, b=t_j}, \\
\left[\mathbf{C}_{\bm{\phi}}' \right]_{i,j} &\coloneqq \frac{\partial}{\partial b} k_{\bm{\phi}}(a, b) \rvert_{a=t_i, b=t_j}, \\
\left[\mathbf{C}_{\bm{\phi}}'' \right]_{i,j} &\coloneqq \frac{\partial^2}{\partial a \partial b} k_{\bm{\phi}}(a, b) \rvert_{a=t_i, b=t_j}.
\end{align}

\subsection{Benchmarking Systems}
\label{sec:AppendixBenchmarkingSystems}
To demonstrate the performance of $\mathrm{ODIN}$, we use four commonly used benchmarking systems. To guarantee a fair comparison, we follow the established parameter settings as adopted among others by \citet{calderhead2009accelerating}, \citet{dondelinger2013ode}, \citet{gorbach2017scalable} and \citet{wenk2018fast}.

\subsubsection{Lotka-Volterra}
The Lotka-Volterra predator-prey model was originally introduced by \citet{lotka1932growth} to describe population dynamics. It is a 2D system whose dynamics are determined by the ODEs
\begin{alignat}{2}
	\dot{x}_1(t) &= \enskip && \theta_1 x_1(t) - \theta_2 x_1(t) x_2(t) \label{eq:LV_true_first} \\
	\dot{x}_2(t) &= \enskip-&& \theta_3 x_2(t) + \theta_4 x_1(t) x_2(t) \label{eq:LV_true_second}.
\end{alignat}
Using $\bm{\theta}=[2, 1, 4, 1]$ and initial conditions $\mathbf{x}(0) = [5, 3]$, the system is in a stable limit cycle with very smooth trajectories. The training data in this case consists of 20 equally spaced observations on the interval $[0, 2]$. It should be noted that the ODEs are linear in one state or parameter variable if all other state and parameter variables are kept constant. In the context of Gaussian process-based gradient matching, this means that the posterior marginals of the parameters and states are Gaussian distributed, which makes this system rather easy to solve. The $\mathrm{VGM}$ algorithm introduced by \citet{gorbach2017scalable} is an excellent showcase of how to use this fact to derive an efficient variational approximation.

\subsubsection{FitzHugh-Nagumo}
The FitzHugh-Nagumo model was originally introduced by \citet{fitzhugh1961impulses} and \citet{nagumo1962active} to model the activation of giant squid neurons. It is a 2D system whose dynamics are described by the ODEs
\begin{align}
	\dot{V} &= \theta_1( V - \frac{V^3}{3} + R)\\
	\dot{R} &= \frac{1}{\theta_1} ( V - \theta_2 + \theta_3 R).
\end{align}
Using $\bm{\theta} = [0.2, 0.2, 3]$ and initial conditions $\mathbf{x}(0) = [-1, 1]$, this system is in a stable limit cycle. However, the trajectories of this system are quite rough with rapidly changing lengthscales, which is a significant challenge for any smoothing-based scheme. Furthermore, both $V$ and $\theta_1$ appear nonlinearly in the ODEs, leading to non-Gaussian posteriors. The dataset for this case consists of 20 equally spaced observations on the interval $[0,10]$.

\subsubsection{Protein Transduction}
The Protein Transduction model was originally introduced by \citet{vyshemirsky2007bayesian} to model chemical reactions in a cell. It is a 5D system whose dynamics are described by the ODEs
\begin{alignat}{2}
	&\dot{S} &&= -\theta_1 S - \theta_2 S R + \theta_3 R_S \nonumber \\
	&\dot{dS} &&= \theta_1 S \nonumber \\
	&\dot{R} &&= -\theta_2 S R + \theta_3 R_S + \theta_5 \frac{R_{pp}}{\theta_6 + R_{pp}} \nonumber \\
	&\dot{R}_S &&= \theta_2 S R - \theta_3 R_S - \theta_4 R_S \nonumber \\
	&\dot{R}_{pp} &&= \theta_4 R_S - \theta_5 \frac{R_{pp}}{\theta_6 + R_{pp}}.
\end{alignat}
The parameters and initial conditions of this system were set respectively to  $\boldsymbol{\theta} = [0.07,0.6,0.05,0.3,0.017,0.3]$ and $\mathbf{x}(0) = [1,0,1,0,0]$. Due to the dynamics changing rapidly at the beginning of the trajectories, the training data is generated by sampling the system at $\mathbf{t} = [0, 1, 2, 4, 5, 7, 10, 15, 20, 30, 40, 50, 60, 80, 100]$. This system has highly nonlinear terms and many claim it to be only weakly identifiable \citep{dondelinger2013ode,gorbach2017scalable,wenk2018fast}.

\subsubsection{Lorenz '96}
The Lorenz '96 system was originally introduced by \citet{lorenz1998optimal} for weather forecasting. The dimensionality $K > 3$ of this model can be chosen arbitrarily, with the $k$-th state being governed by the differential equation
\begin{equation}
	\dot{x}_k = (x_{k+1} - x_{k-2})x_{k-1} - x_k + \theta,
\end{equation}
where all indices should be read modulo $K$. Introduced to the Gaussian process-based gradient matching community by \citet{gorbach2017scalable}, this system is used to demonstrate the scaling of any algorithm in the amount of states $K$, by keeping everything but $K$ fixed. Following \citet{gorbach2017scalable}, we use $\theta=8$ and 50 equally spaced observations over $t=[0, 5]$.

\subsection{State Inference}
\label{sec:AppendixStates}
\subsubsection{Problem Setting}
Let us assume we are provided with a set of noisy observations $\mathbf{y}$ measured at the times $\mathbf{t}$ and the true parametric form $\dot{\mathbf{x}} = f(\mathbf{x}, \bm{\theta})$. The objective consists in recovering the true states $\mathbf{x}^*$ at observation times $\mathbf{t}$ without any information about the true parameters $\bm{\theta}^*$.

\subsubsection{ODEs Provide Useful Information}
\begin{figure}
	\centering
	\begin{subfigure}[t]{0.325\textwidth}
		\centering
		\includegraphics{plots/states_estimation/boxplots_odin_gpr/odin_gpr_states_mse_lv_0p1_noise.pdf}
	\end{subfigure}
	\hfill
	\begin{subfigure}[t]{0.325\textwidth}
		\centering
		\includegraphics{plots/states_estimation/boxplots_odin_gpr/odin_gpr_states_mse_fhn_100_snr.pdf}
	\end{subfigure}
	\hfill
	\begin{subfigure}[t]{0.325\textwidth}
		\centering
		\includegraphics{plots/states_estimation/boxplots_odin_gpr/odin_gpr_states_mse_pt_0p001_noise.pdf}
	\end{subfigure}\\
	\begin{subfigure}[t]{0.325\textwidth}
		\centering
		\includegraphics{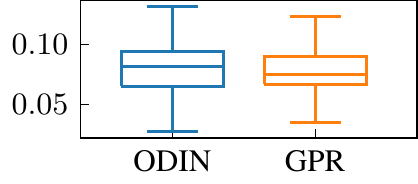}
		\caption{Lotka-Volterra}
	\end{subfigure}
	\hfill
	\begin{subfigure}[t]{0.325\textwidth}
		\centering
		\includegraphics{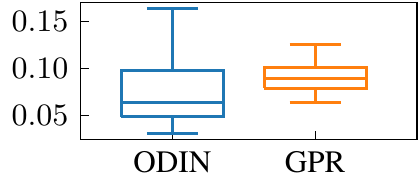}
		\caption{FitzHugh-Nagumo}
	\end{subfigure}
	\hfill
	\begin{subfigure}[t]{0.325\textwidth}
		\centering
		\includegraphics{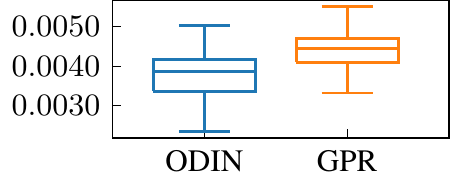}
		\caption{Protein Transduction}
	\end{subfigure}
	\caption{Comparing the RMSE of state estimates using vanilla GP regression and $\mathrm{ODIN}$. All systems were evaluated on 100 independent noise realizations and parameter initializations. The top row shows the low noise case, the bottom row shows the high noise case.}
	\label{fig:stateRMSEAppendix}
\end{figure}

As shown in Figure \ref{fig:stateRMSEAppendix}, $\mathrm{ODIN}$ provides more accurate state estimates in all but the high noise LV case compared to standard GP regression. The different behavior can be explained by the quality of the GP prior. For Lotka Volterra, the RBF kernel provides a perfectly suited prior for the sinusoidal form of its dynamics. It is thus not surprising that the GP regression estimates are already quite good, especially in a high noise setting. However, for both FitzHugh Nagumo and Protein Transduction, the GP prior is slightly off. Thus, including the additional information provided by the ODEs leads to significant improvements in state estimation.

\ \\ \ \\ \ \\ \ \\ \ \\ \ \\ \ \\ \ \\ \ \\ \ \\ \ \\ \ \\ \ \\ \ \\ \ \\ \ \\ \ \\

\subsection{Median Trajectories}
\label{sec:AppendixMedianTrajectories}

\vfill

\begin{figure}[h]
	\centering
	\begin{subfigure}[t]{0.245\textwidth}
		\centering
		\includegraphics{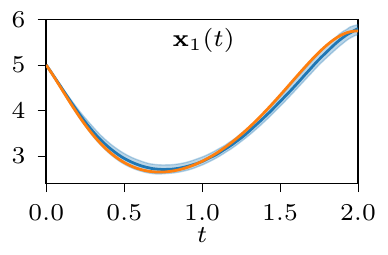}
	\end{subfigure}
	\hfill
	\begin{subfigure}[t]{0.245\textwidth}
		\centering
		\includegraphics{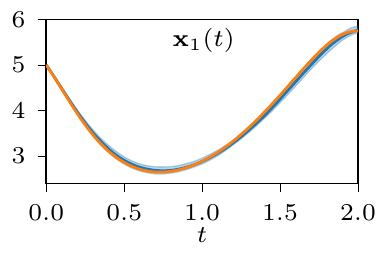}
	\end{subfigure}
	\hfill
	\begin{subfigure}[t]{0.245\textwidth}
		\centering
		\includegraphics{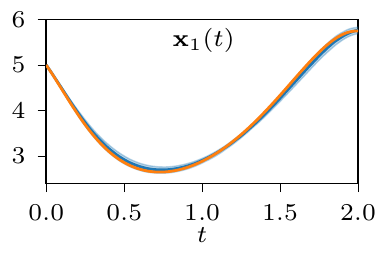}
	\end{subfigure}
	\hfill
	\begin{subfigure}[t]{0.245\textwidth}
		\centering
		\includegraphics{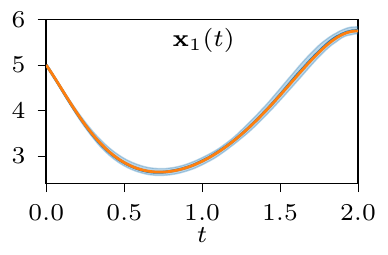}
	\end{subfigure}\\
	\begin{subfigure}[t]{0.245\textwidth}
		\centering
		\includegraphics{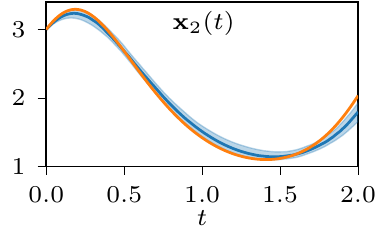}
		\caption{$\mathrm{AGM}$}
	\end{subfigure}
	\hfill
	\begin{subfigure}[t]{0.245\textwidth}
		\centering
		\includegraphics{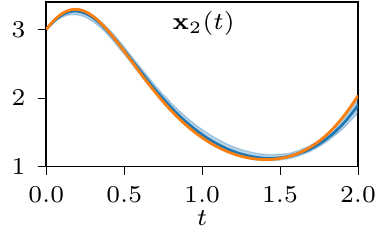}
		\caption{$\mathrm{FGPGM}$}
	\end{subfigure}
	\hfill
	\begin{subfigure}[t]{0.245\textwidth}
		\centering
		\includegraphics{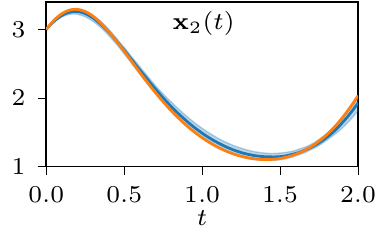}
		\caption{$\mathrm{ODIN}$}
	\end{subfigure}
	\hfill
	\begin{subfigure}[t]{0.245\textwidth}
		\centering
		\includegraphics{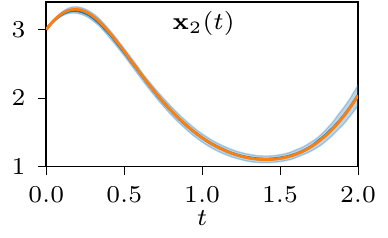}
		\caption{$\mathrm{RKG3}$}
	\end{subfigure}
	\caption{Comparison of the trajectories obtained by numerically integrating the inferred parameters of the Lotka-Volterra system for $\sigma=0.1$. The plot was created using 100 independent noise realizations, where the solid blue line is the median trajectory and the shaded areas denote the 25\% and 75\% quantiles. While the upper row shows the results for state $x_1$, the lower one does the same for state $x_2$. The orange trajectory is the ground truth. The difference between the four algorithms is barely visible.}
\end{figure}

\vfill

\begin{figure}[h]
	\centering
	\begin{subfigure}[t]{0.245\textwidth}
		\centering
		\includegraphics{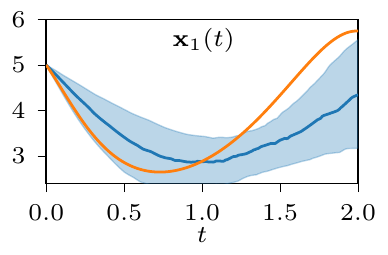}
	\end{subfigure}
	\hfill
	\begin{subfigure}[t]{0.245\textwidth}
		\centering
		\includegraphics{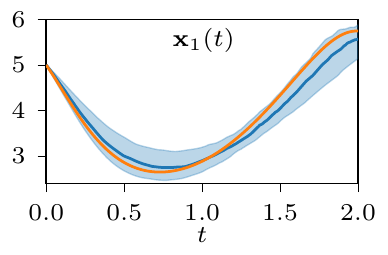}
	\end{subfigure}
	\hfill
	\begin{subfigure}[t]{0.245\textwidth}
		\centering
		\includegraphics{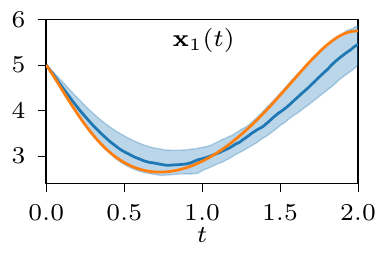}
	\end{subfigure}
	\hfill
	\begin{subfigure}[t]{0.245\textwidth}
		\centering
		\includegraphics{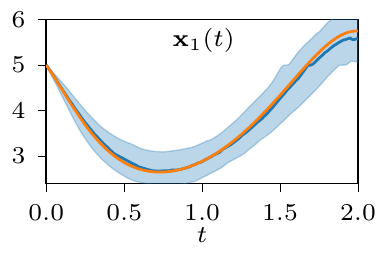}
	\end{subfigure}\\
	\begin{subfigure}[t]{0.245\textwidth}
		\centering
		\includegraphics{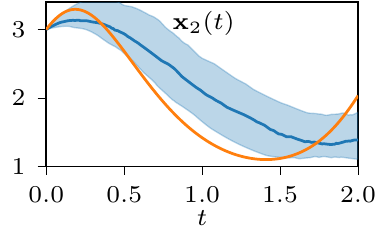}
		\caption{$\mathrm{AGM}$}
	\end{subfigure}
	\hfill
	\begin{subfigure}[t]{0.245\textwidth}
		\centering
		\includegraphics{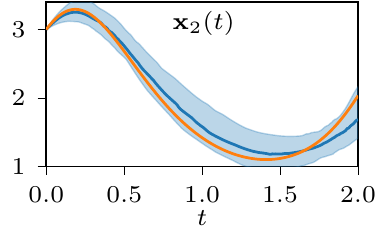}
		\caption{$\mathrm{FGPGM}$}
	\end{subfigure}
	\hfill
	\begin{subfigure}[t]{0.245\textwidth}
		\centering
		\includegraphics{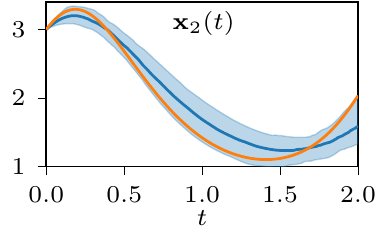}
		\caption{$\mathrm{ODIN}$}
	\end{subfigure}
	\hfill
	\begin{subfigure}[t]{0.245\textwidth}
		\centering
		\includegraphics{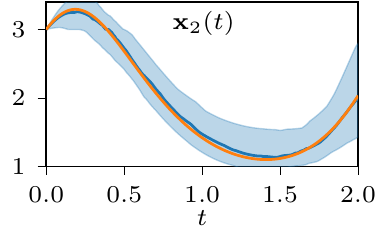}
		\caption{$\mathrm{RKG3}$}
	\end{subfigure}
	\caption{Comparison of the trajectories obtained by numerically integrating the inferred parameters of the Lotka-Volterra system for $\sigma=0.5$. The plot was created using 100 independent noise realizations, where the solid blue line is the median trajectory and the shaded areas denote the 25\% and 75\% quantiles. While the upper row shows the results for state $x_1$, the lower one does the same for state $x_2$. The orange trajectory is the ground truth. While all algorithms seem to perform reasonably well, the perfect match between the sinusoidal dynamics and the RBF kernel lead to a well performing RKG3, while the more flexible Gaussian process based schemes seem to suffer more strongly from a smoothing bias.}
\end{figure}

\vfill

\vfill

\begin{figure}[h]
	\centering
	\begin{subfigure}[t]{0.325\textwidth}
		\centering
		\includegraphics{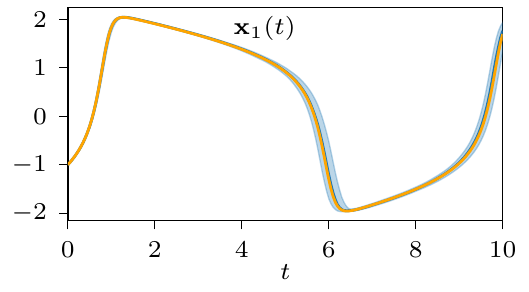}
	\end{subfigure}
	\hfill
	\begin{subfigure}[t]{0.325\textwidth}
		\centering
		\includegraphics{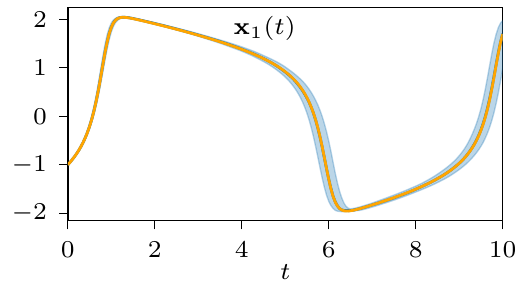}
	\end{subfigure}
	\hfill
	\begin{subfigure}[t]{0.325\textwidth}
		\centering
		\includegraphics{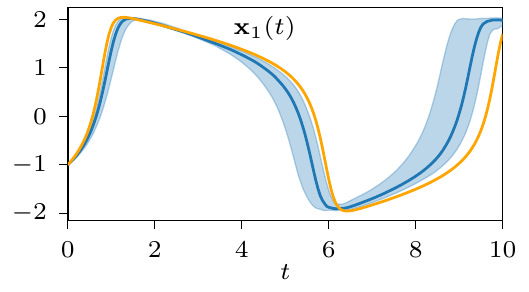}
	\end{subfigure}\\
	\begin{subfigure}[t]{0.325\textwidth}
		\centering
		\includegraphics{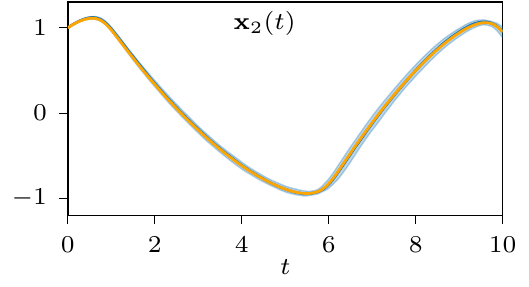}
		\caption{$\mathrm{FGPGM}$}
	\end{subfigure}
	\hfill
	\begin{subfigure}[t]{0.325\textwidth}
		\centering
		\includegraphics{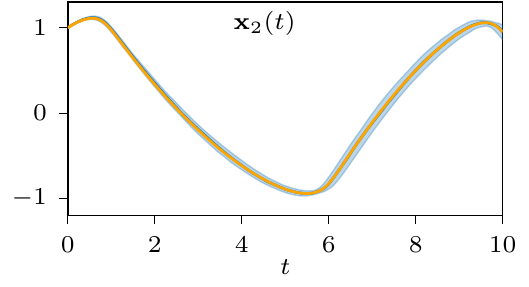}
		\caption{$\mathrm{ODIN}$}
	\end{subfigure}
	\hfill
	\begin{subfigure}[t]{0.325\textwidth}
		\centering
		\includegraphics{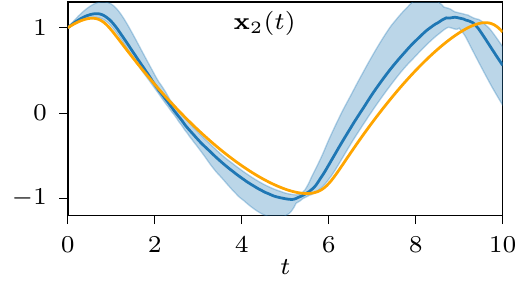}
		\caption{$\mathrm{RKG3}$}
	\end{subfigure}
	\caption{Comparison of the trajectories obtained by numerically integrating the inferred parameters of the FitzHugh-Nagumo system for a SNR of 100. The plot was created using 100 independent noise realizations, where the solid blue line is the median trajectory and the shaded areas denote the 25\% and 75\% quantiles. While the upper row shows the results for state $x_1$, the lower one does the same for state $x_2$. The orange trajectory indicates the ground truth. This experiment demonstrates how $\mathrm{ODIN}$ can comfortably deal with dynamics with rapidly changing lengthscales.}
\end{figure}
\vfill
\begin{figure}[h]
	\centering
	\begin{subfigure}[t]{0.325\textwidth}
		\centering
		\includegraphics{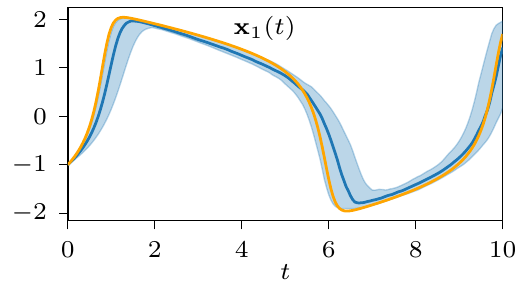}
	\end{subfigure}
	\hfill
	\begin{subfigure}[t]{0.325\textwidth}
		\centering
		\includegraphics{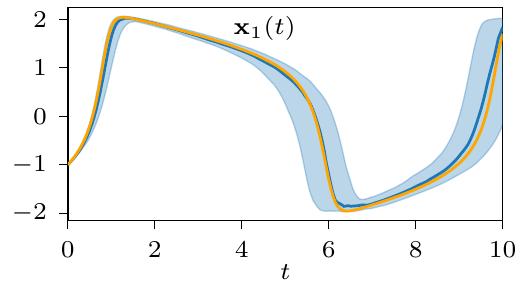}
	\end{subfigure}
	\hfill
	\begin{subfigure}[t]{0.325\textwidth}
		\centering
		\includegraphics{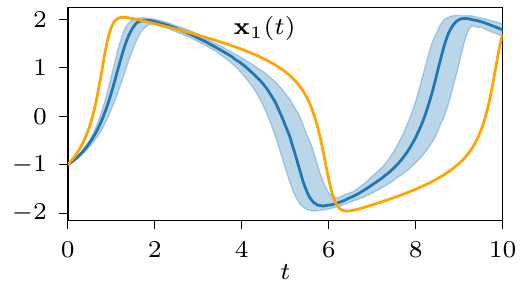}
	\end{subfigure}\\
	\begin{subfigure}[t]{0.325\textwidth}
		\centering
		\includegraphics{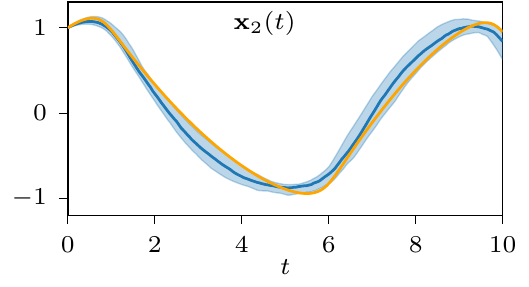}
		\caption{$\mathrm{FGPGM}$}
	\end{subfigure}
	\hfill
	\begin{subfigure}[t]{0.325\textwidth}
		\centering
		\includegraphics{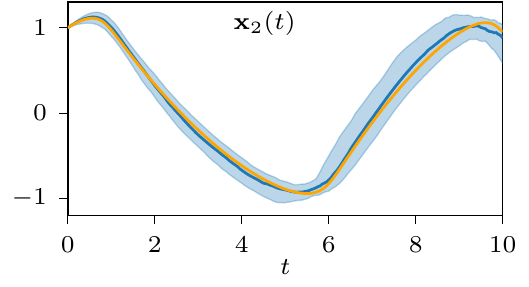}
		\caption{$\mathrm{ODIN}$}
	\end{subfigure}
	\hfill
	\begin{subfigure}[t]{0.325\textwidth}
		\centering
		\includegraphics{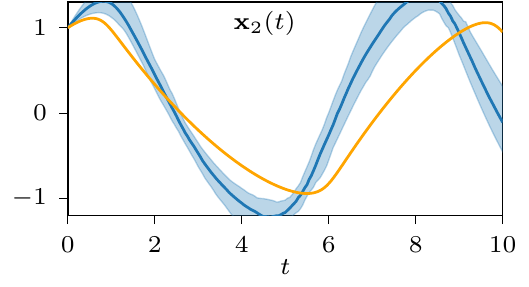}
		\caption{$\mathrm{RKG3}$}
	\end{subfigure}
	\caption{Comparison of the trajectories obtained by numerically integrating the inferred parameters of the FitzHugh-Nagumo system for a SNR of 10. The plot was created using 100 independent noise realizations, where the solid blue line is the median trajectory and the shaded areas denote the 25\% and 75\% quantiles. While the upper row shows the results for state $x_1$, the lower one does the same for state $x_2$. The orange trajectory indicates the ground truth. This experiment demonstrates how $\mathrm{ODIN}$ can comfortably deal with dynamics with rapidly changing lengthscales.}
\end{figure}

\vfill


\vfill

\begin{figure}[h]
	\centering
	\begin{subfigure}[t]{0.245\textwidth}
		\centering
		\includegraphics{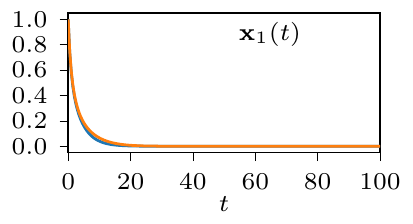}
	\end{subfigure}
	\hfill
	\begin{subfigure}[t]{0.245\textwidth}
		\centering
		\includegraphics{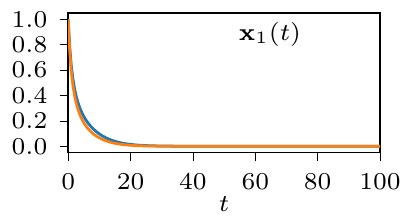}
	\end{subfigure}
	\hfill
	\begin{subfigure}[t]{0.245\textwidth}
		\centering
		\includegraphics{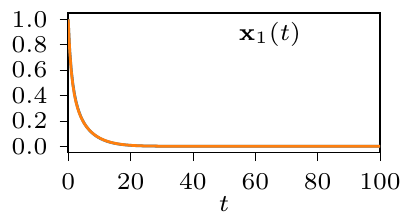}
	\end{subfigure}
	\hfill
	\begin{subfigure}[t]{0.245\textwidth}
		\centering
		\includegraphics{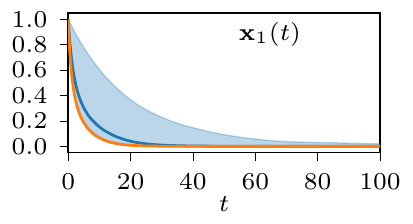}
	\end{subfigure}\\
	\begin{subfigure}[t]{0.245\textwidth}
		\centering
		\includegraphics{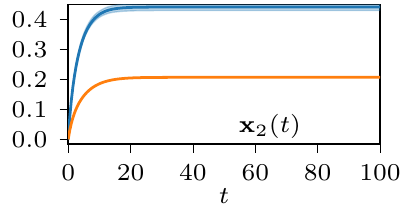}
	\end{subfigure}
	\hfill
	\begin{subfigure}[t]{0.245\textwidth}
		\centering
		\includegraphics{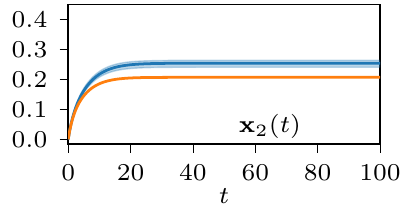}
	\end{subfigure}
	\hfill
	\begin{subfigure}[t]{0.245\textwidth}
		\centering
		\includegraphics{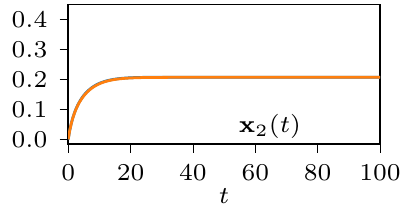}
	\end{subfigure}
	\hfill
	\begin{subfigure}[t]{0.245\textwidth}
		\centering
		\includegraphics{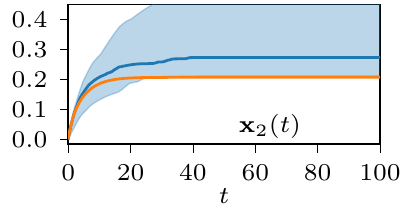}
	\end{subfigure}\\
	\begin{subfigure}[t]{0.245\textwidth}
		\centering
		\includegraphics{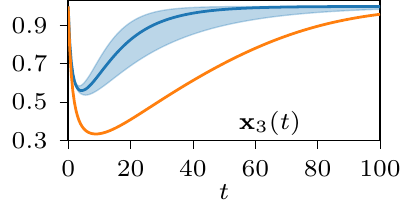}
	\end{subfigure}
	\hfill
	\begin{subfigure}[t]{0.245\textwidth}
		\centering
		\includegraphics{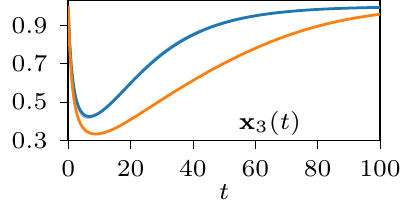}
	\end{subfigure}
	\hfill
	\begin{subfigure}[t]{0.245\textwidth}
		\centering
		\includegraphics{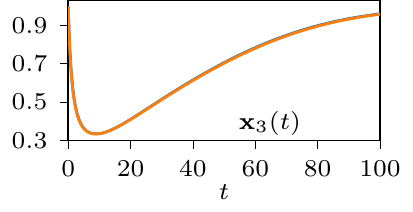}
	\end{subfigure}
	\hfill
	\begin{subfigure}[t]{0.245\textwidth}
		\centering
		\includegraphics{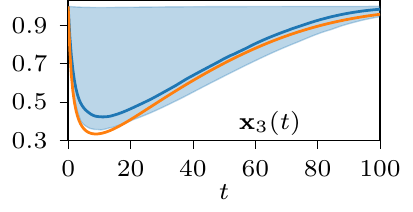}
	\end{subfigure}\\
	\begin{subfigure}[t]{0.245\textwidth}
		\centering
		\includegraphics{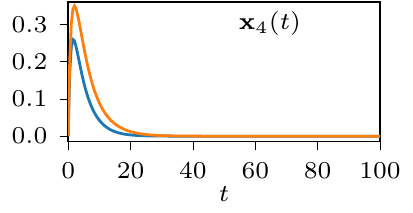}
	\end{subfigure}
	\hfill
	\begin{subfigure}[t]{0.245\textwidth}
		\centering
		\includegraphics{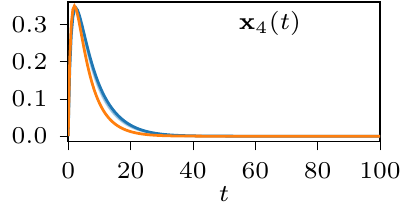}
	\end{subfigure}
	\hfill
	\begin{subfigure}[t]{0.245\textwidth}
		\centering
		\includegraphics{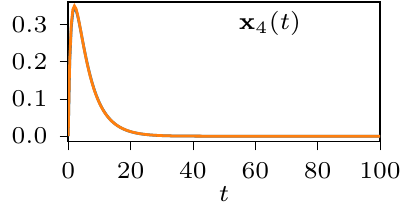}
	\end{subfigure}
	\hfill
	\begin{subfigure}[t]{0.245\textwidth}
		\centering
		\includegraphics{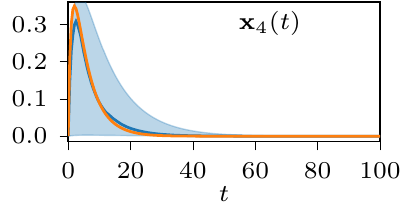}
	\end{subfigure}\\
	\begin{subfigure}[t]{0.245\textwidth}
		\centering
		\includegraphics{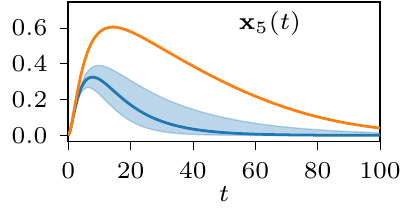}
		\caption{$\mathrm{AGM}$}
	\end{subfigure}
	\hfill
	\begin{subfigure}[t]{0.245\textwidth}
		\centering
		\includegraphics{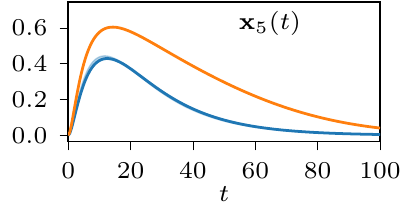}
		\caption{$\mathrm{FGPGM}$}
	\end{subfigure}
	\hfill
	\begin{subfigure}[t]{0.245\textwidth}
		\centering
		\includegraphics{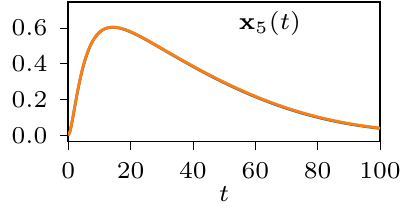}
		\caption{$\mathrm{ODIN}$}
	\end{subfigure}
	\hfill
	\begin{subfigure}[t]{0.245\textwidth}
		\centering
		\includegraphics{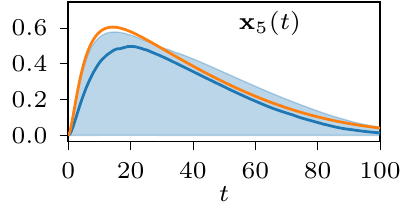}
		\caption{$\mathrm{RKG3}$}
	\end{subfigure}
	\caption{Comparison of the trajectories obtained by numerically integrating the inferred parameters of the Protein Transduction system for $\sigma=0.001$. The plot was created using 100 independent noise realizations, where the solid blue line is the median trajectory and the shaded areas denote the 23\% and 75\% quantiles. The orange trajectory is the ground truth. This experiment proves how $\mathrm{ODIN}$ is the only algorithm among the displayed ones that can handle non-Gaussian posterior marginals.}
\end{figure}

\begin{figure}[h]
	\centering
	\begin{subfigure}[t]{0.245\textwidth}
		\centering
		\includegraphics{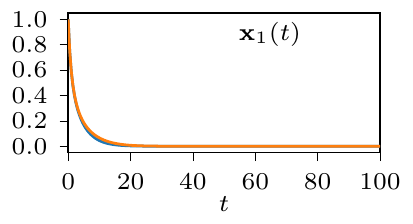}
	\end{subfigure}
	\hfill
	\begin{subfigure}[t]{0.245\textwidth}
		\centering
		\includegraphics{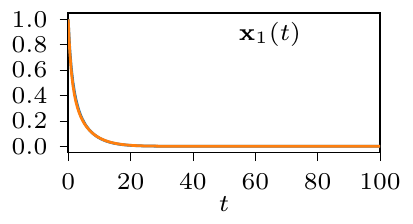}
	\end{subfigure}
	\hfill
	\begin{subfigure}[t]{0.245\textwidth}
		\centering
		\includegraphics{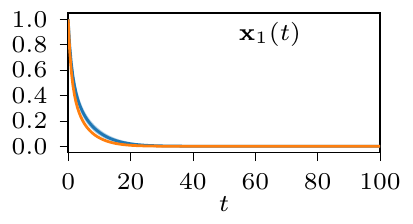}
	\end{subfigure}
	\hfill
	\begin{subfigure}[t]{0.245\textwidth}
		\centering
		\includegraphics{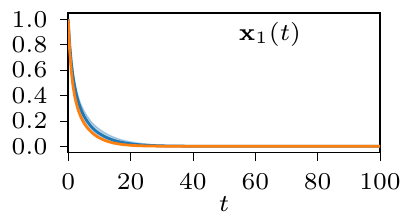}
	\end{subfigure}\\
	\begin{subfigure}[t]{0.245\textwidth}
		\centering
		\includegraphics{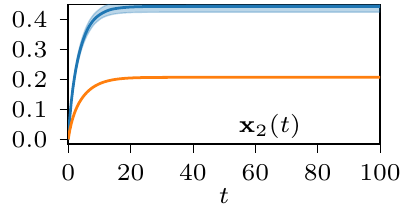}
	\end{subfigure}
	\hfill
	\begin{subfigure}[t]{0.245\textwidth}
		\centering
		\includegraphics{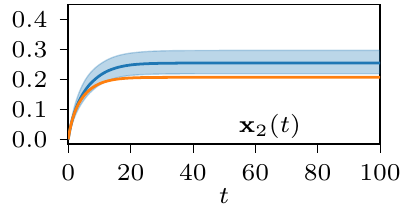}
	\end{subfigure}
	\hfill
	\begin{subfigure}[t]{0.245\textwidth}
		\centering
		\includegraphics{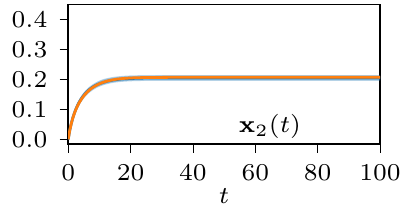}
	\end{subfigure}
	\hfill
	\begin{subfigure}[t]{0.245\textwidth}
		\centering
		\includegraphics{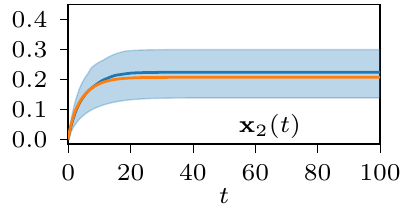}
	\end{subfigure}\\
	\begin{subfigure}[t]{0.245\textwidth}
		\centering
		\includegraphics{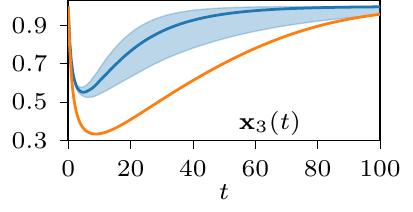}
	\end{subfigure}
	\hfill
	\begin{subfigure}[t]{0.245\textwidth}
		\centering
		\includegraphics{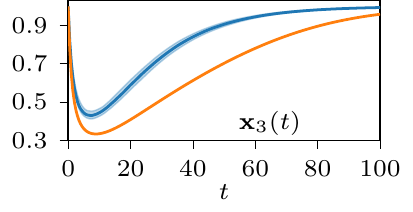}
	\end{subfigure}
	\hfill
	\begin{subfigure}[t]{0.245\textwidth}
		\centering
		\includegraphics{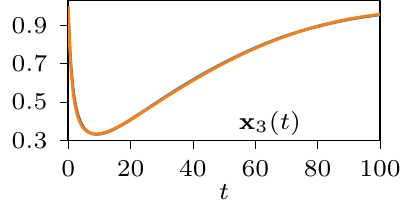}
	\end{subfigure}
	\hfill
	\begin{subfigure}[t]{0.245\textwidth}
		\centering
		\includegraphics{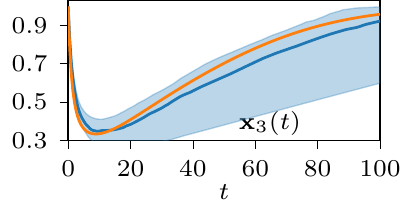}
	\end{subfigure}\\
	\begin{subfigure}[t]{0.245\textwidth}
		\centering
		\includegraphics{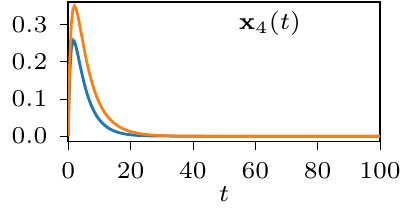}
	\end{subfigure}
	\hfill
	\begin{subfigure}[t]{0.245\textwidth}
		\centering
		\includegraphics{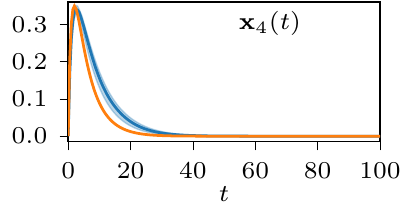}
	\end{subfigure}
	\hfill
	\begin{subfigure}[t]{0.245\textwidth}
		\centering
		\includegraphics{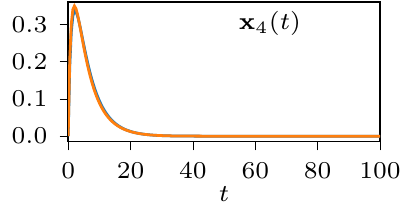}
	\end{subfigure}
	\hfill
	\begin{subfigure}[t]{0.245\textwidth}
		\centering
		\includegraphics{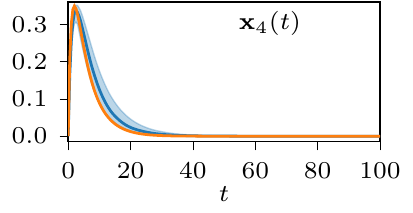}
	\end{subfigure}\\
	\begin{subfigure}[t]{0.245\textwidth}
		\centering
		\includegraphics{plots/medians/pt/agm/noise_0p01/state4.pdf}
		\caption{$\mathrm{AGM}$}
	\end{subfigure}
	\hfill
	\begin{subfigure}[t]{0.245\textwidth}
		\centering
		\includegraphics{plots/medians/pt/fgpgm/noise_0p01/state4.pdf}
		\caption{$\mathrm{FGPGM}$}
	\end{subfigure}
	\hfill
	\begin{subfigure}[t]{0.245\textwidth}
		\centering
		\includegraphics{plots/medians/pt/odin/noise_0p01/state4.pdf}
		\caption{$\mathrm{ODIN}$}
	\end{subfigure}
	\hfill
	\begin{subfigure}[t]{0.245\textwidth}
		\centering
		\includegraphics{plots/medians/pt/niu/noise_0p01/state4.pdf}
		\caption{$\mathrm{RKG3}$}
	\end{subfigure}
	\caption{Comparison of the trajectories obtained by numerically integrating the inferred parameters of the Protein Transduction system for $\sigma=0.01$. The plot was created using 100 independent noise realizations, where the solid blue line is the median trajectory and the shaded areas denote the 23\% and 75\% quantiles. The orange trajectory is the ground truth. As in the low noise case, this experiment proves how $\mathrm{ODIN}$ is the only algorithm among the displayed ones that can handle non-Gaussian posterior marginals.}
\end{figure}

\begin{figure}[h]
	\large \textbf{Statewise tRMSE}\par\medskip \ \\ \ \\ 
	\centering
	\begin{subfigure}[t]{0.475\textwidth}
		\centering
		\includegraphics{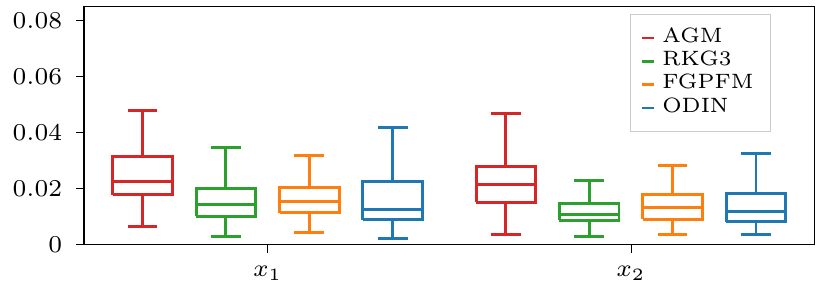}
	\end{subfigure}
	\hfill
	\begin{subfigure}[t]{0.475\textwidth}
		\centering
		\includegraphics{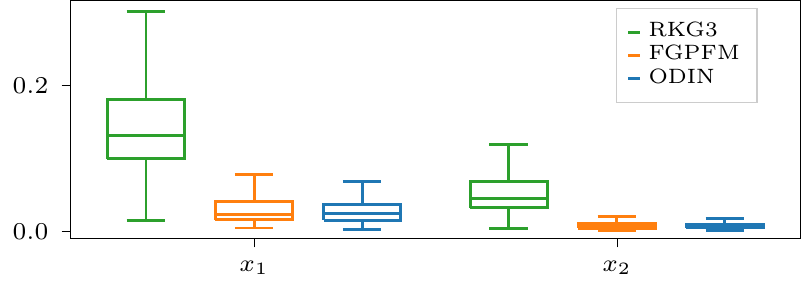}
	\end{subfigure}\\
	\begin{subfigure}[t]{0.475\textwidth}
		\centering
		\includegraphics{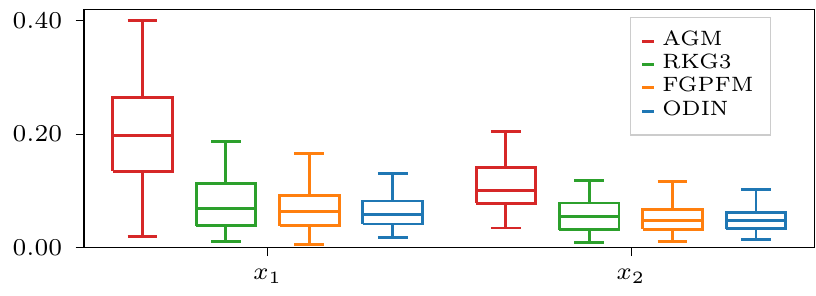}
		\caption{Lotka-Volterra}
	\end{subfigure}
	\hfill
	\begin{subfigure}[t]{0.475\textwidth}
		\centering
		\includegraphics{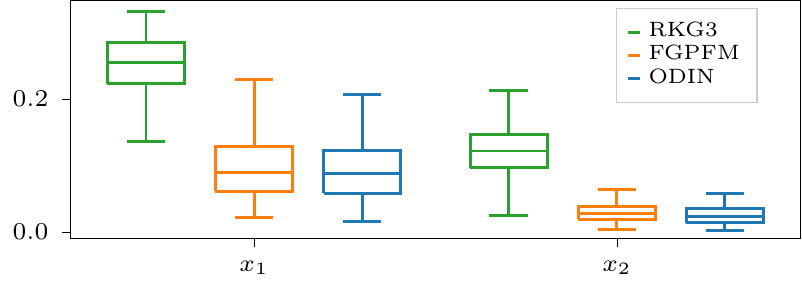}
		\caption{FitzHugh-Nagumo}
	\end{subfigure}\\
	\begin{subfigure}[t]{\textwidth}
		\centering
		\includegraphics{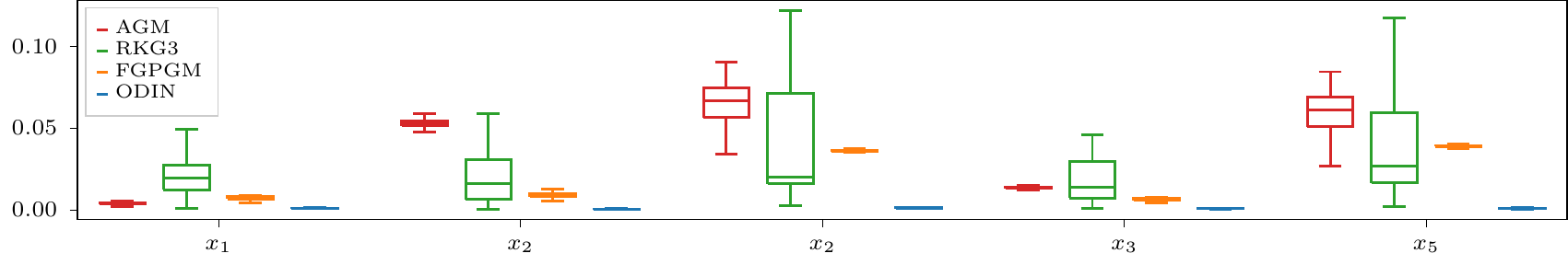}
	\end{subfigure}\\
	\begin{subfigure}[t]{\textwidth}
		\centering
		\includegraphics{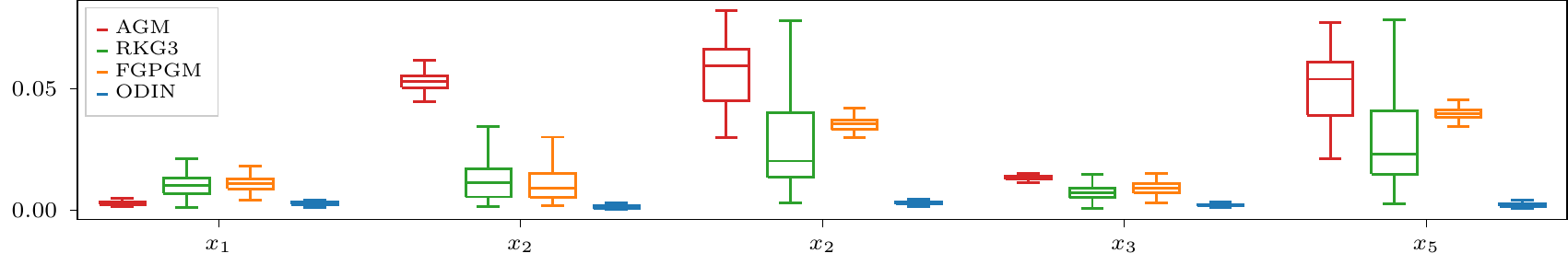}
		\caption{Protein Transduction}
	\end{subfigure}\\
	\caption{Statewise trajectory RMSE for all benchmark systems in the parameter inference problem. For each pair of plots, the top shows the low noise case with $\sigma=0.1$ for LV, $\sigma=0.001$ for PT and $SNR=100$ for FHN. The bottom shows the high noise case with $\sigma=0.5$ for LV, $\sigma=0.01$ for PT and $SNR=10$ for FHN.}
	\label{fig:statewise_tRMSE}
	\ \\ \ \\ \ \\ \ \\ \ \\ \ \\ \ \\ \ \\ \ \\ \ \\ \ \\ \ \\ \ \\ \ \\
\end{figure}

\vfill